\documentclass{article}

\PassOptionsToPackage{round}{natbib}
\usepackage[final]{neurips_data_2022}

\usepackage[utf8]{inputenc} 
\usepackage[T1]{fontenc}    
\usepackage{hyperref}       
\usepackage{url}            
\usepackage{booktabs}       
\usepackage{amsfonts}       
\usepackage{nicefrac}       
\usepackage{microtype}      
\usepackage{xcolor}         
\usepackage{todonotes}
\usepackage{amsmath}
\usepackage{caption}
\usepackage{subcaption}
\usepackage{wrapfig}
\usepackage{dsfont}
\usepackage{nccmath}
\usepackage[ruled,vlined]{algorithm2e}
\usepackage{multirow}
\usepackage{graphicx}
\usepackage{wrapfig}
\usepackage{listings}
\title{DDXPlus: A New Dataset For Automatic Medical Diagnosis}

\newcommand*\samethanks[1][\value{footnote}]{\footnotemark[#1]}

\author{
    Arsène Fansi Tchango \thanks{Equal contribution} $^{ , \ddagger}$, Rishab Goel \samethanks[1] $^{ , \dagger}$, Zhi Wen $^{\ddagger}$, Julien Martel $^{\S}$, Joumana Ghosn $^{\ddagger}$ \\
  $^{\ddagger}$ Mila-Quebec AI Institute \\
  \texttt{\{arsene.fansi.tchango,zhi.wen,joumana.ghosn\}@mila.quebec} \\
  $^{\dagger}$ \texttt{rgoel0112@gmail.com} \\
  $^{\S}$ Work done while at Dialogue Health Technologies Inc. \\
  \texttt{julien@datadoc.ca} \\
}

\newcommand{\Symcat}{\textit{SymCAT} }
\newcommand{\Synthea}{\textit{Synthea\textsuperscript{TM}} }

\newcommand{\DXA}{DXA}

\begin{document}

\maketitle

\begin{abstract}
  There has been a rapidly growing interest in Automatic Symptom Detection (ASD) and Automatic Diagnosis (AD) systems in the machine learning research literature, aiming to assist doctors in telemedicine services. These systems are designed to interact with patients, collect evidence about their symptoms and relevant antecedents, and possibly make predictions about the underlying diseases. Doctors would review the interactions, including the evidence and the predictions, collect if necessary additional information from patients, before deciding on next steps. Despite recent progress in this area, an important piece of doctors' interactions with patients is missing in the design of these systems, namely the differential diagnosis. Its absence is largely due to the lack of datasets that include such information for models to train on. In this work, we present a large-scale synthetic dataset of roughly 1.3 million patients that includes a differential diagnosis, along with the ground truth pathology, symptoms and antecedents for each patient. Unlike existing datasets which only contain binary symptoms and antecedents, this dataset also contains categorical and multi-choice symptoms and antecedents useful for efficient data collection. Moreover, some symptoms are organized in a hierarchy, making it possible to design systems able to interact with patients in a logical way. As a proof-of-concept, we extend two existing AD and ASD systems to incorporate the differential diagnosis, and provide empirical evidence that using differentials as training signals is essential for the efficiency of such systems or for helping doctors better understand the reasoning of those systems. The dataset is available at \href{https://figshare.com/articles/dataset/DDXPlus_Dataset/20043374}{https://figshare.com/articles/dataset/DDXPlus\_Dataset/20043374}. 
\end{abstract}

\section{Introduction}
In a clinical conversation between a doctor and a patient, the patient usually initiates the discussion by specifying an initial set of symptoms they are experiencing. The doctor then iteratively inquires about additional symptoms and antecedents (describing the patient’s medical history and potential risk factors), and considers, throughout the interaction, a differential diagnosis, i.e., a short list of plausible diseases the patient might be suffering from \citep{henderson2012patient,guyatt2002users,rhoads2017formulating}, which is refined based on the patient's responses. 
During this multi-step process, the doctor tries to collect all relevant information to narrow down the differential diagnosis. Once the differential diagnosis is finalized, the doctor can ask the patient to undergo medical exams to eliminate most pathologies included in it and confirm the one(s) the patient is suffering from, or can decide to directly prescribe a treatment to the patient. 

Aiming to aid doctors in such clinical interactions, there has recently been significant progress in building Automatic Symptom Detection (ASD) and Automatic Diagnosis (AD) systems, using machine learning and Reinforcement Learning (RL) techniques \citep{wei2018task,xu2019end,chen2021diaformer,zhao2021weighted,guan2021bayesian,yuan2021efficient,liu2022my}. These systems are meant to collect symptoms and antecedents relevant to the patient's condition, while minimizing the length of the interaction to improve efficiency and avoid burdening the patient with unnecessary questions. In addition, AD systems are tasked to predict the patient's underlying disease to further aid doctors in deciding appropriate next steps. However, this setting differs from real patient-doctor interactions in an important way, namely the absence of the differential diagnosis. Based on the conversation alone, without further information such as the results of medical exams, doctors tend to consider in their reasoning a differential diagnosis rather than a single pathology \citep{henderson2012patient}. In addition to presenting a more comprehensive view of the doctor's opinion on the patient's underlying condition, the differential diagnosis helps account for the uncertainty in the diagnosis, since a patient's antecedents and symptoms can point to multiple pathologies. Moreover, the differential diagnosis can help guide the doctor in determining which questions to ask the patient during their interaction. Thus, considering the differential diagnosis is especially important for better and more efficient evidence collection, for accounting for the uncertainty in the diagnosis, and for building systems that doctors can understand and trust. We believe that the absence of this key ingredient in recent AD/ASD systems is mainly due to the lack of datasets that include such information. The most commonly used public datasets, such as DX \citep{wei2018task}, Muzhi \citep{xu2019end} and SymCAT \citep{peng2018refuel}, are all designed for predicting the pathology a patient is experiencing and do not contain differential diagnosis data.

To close this gap and encourage future research that focuses on the differential diagnosis, we introduce DDXPlus, a large-scale synthetic dataset for building AD and ASD systems. This dataset is similar in format to other public datasets such as DX \citep{wei2018task} and Muzhi \citep{xu2019end}, but differs in several important ways. First, it makes a clear distinction between symptoms and antecedents which are not of equal importance from a doctor's perspective when interacting with patients. Second, it is larger in scale, in terms of the number of patients, as well as the number of represented pathologies, symptoms and antecedents. Third, contrary to existing datasets which only include binary symptoms and antecedents, it also includes categorical and multi-choice symptoms and antecedents useful for efficient evidence collection. Moreover, some symptoms are organized in a hierarchy making it possible to design systems able to interact with patients in a logical way. Finally, each patient is characterized by a differential diagnosis as well as the pathology they are actually suffering from. To the best of our knowledge, this is the first large-scale dataset that includes both ground truth pathologies and differential diagnoses, as well as non-binary symptoms and antecedents. To summarize, we make the following contributions:

\begin{itemize}
\item We release a large-scale synthetic benchmark dataset of roughly 1.3 million patients covering 49 pathologies, 110 symptoms and 113 antecedents. The dataset is generated using a proprietary medical knowledge base and a commercial AD system, and contains a mixture of multi-choice, categorical and binary symptoms and antecedents. Importantly, it also contains a differential diagnosis for each patient along with the patient's underlying pathology.
\item We extend two existing AD and ASD systems to incorporate the differential diagnosis and show that using this information as a training signal improves their performance or helps doctors understand their reasoning by comparing the collected evidence and the differential.
\end{itemize}

\section{Existing datasets and limitations}
To build machine learning-based AD or ASD systems that medical doctors can trust, one needs to have access to related patient data, namely, the whole set of symptoms experienced by the patient, the relevant antecedents, the underlying pathology, and lastly the differential diagnosis associated with the experienced symptoms and antecedents.
Unfortunately, there is no public dataset with all these characteristics. Existing public datasets of medical records, such as the MIMIC-III dataset \citep{mimiciii}, often lack symptom-related data and are therefore inappropriate for training AD/ASD models. Other datasets, such as DX \citep{wei2018task} or Muzhi \citep{xu2019end}, are of small scale, and don't necessarily provide a holistic view of the symptoms and antecedents experienced by patients since they are derived from medical conversations. Indeed, if a symptom is not mentioned in a conversation, there is no way to determine if it was experienced by the underlying patient or not.

To tackle these limitations, previous works \citep{peng2018refuel, kao2018context} relied on the SymCAT database \citep{Symcat} for data synthesis. Unfortunately, SymCAT includes binary-only symptoms, which can lead to unnecessarily long interactions with patients, compared to categorical or multi-choice questions that collect the same information in fewer turns. For example, to ask about pain location, systems built using SymCAT need to ask several questions, such  as \textit{Do you have pain in your arm?}, \textit{Do you have pain in your ankle?}, $\cdots$, until every relevant location is inquired about. The same information can be obtained by asking a more generic question such as \textit{Where is your pain located?} and obtain all the locations at once. Moreover, as noted by \citep{yuan2021efficient}, the symptom information in SymCAT is incomplete and, as a consequence, the synthetic patients generated using SymCAT are not sufficiently realistic for testing AD and ASD systems. Other datasets with more realistic synthetic patients have recently been proposed based on the Human Phenotype Ontology or HPO \citep{kohler2021human} and MedlinePlus \citep{yuan2021efficient}. However, like SymCAT, they contain only binary symptoms. 

Most importantly, a critical shortcoming that is common to all aforementioned datasets is they do not contain differential diagnosis data. In what follows, we introduce the DDXPlus dataset and describe the undertaken steps towards its creation.

\section{DDXPlus dataset} 
\label{sec:ds}

\begin{figure}
    \centering
    \includegraphics[width=0.8\textwidth]{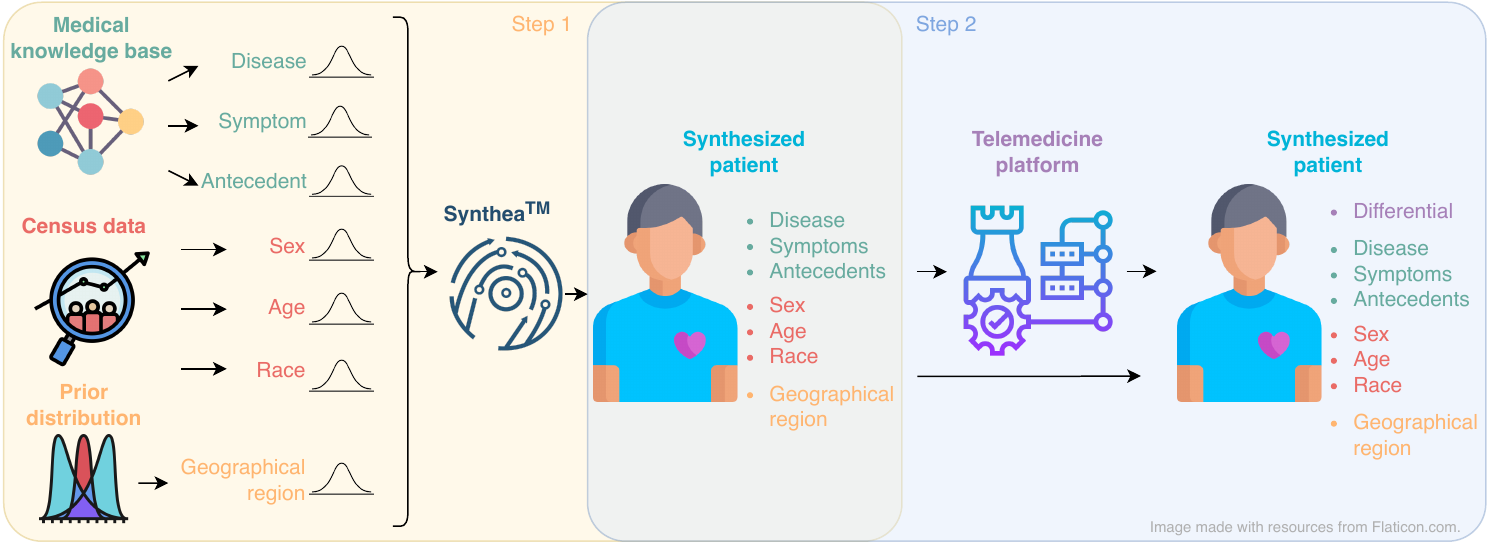}
    \caption{Overview of the data generation process of DDXPlus.}
    \label{fig:overview}
\end{figure}

To generate the dataset, we proceed in two steps. For each patient, we first use a proprietary medical knowledge base (Section \ref{ss:kb}), public census data, and \Synthea \citep{Walonoski2017Synthea} to synthesize their socio-demographic data, underlying disease, symptoms and antecedents (Section \ref{ss:patient-generation}). Next, we generate the patient's corresponding differential diagnosis using an existing commercial rule-based AD system (Section \ref{ss:diff-generation})\footnote{Both the knowledge base and the rule-based AD system used throughout this work are provided by Dialogue Health Technologies Inc.  More information about both components is provided in Appendix \ref{appx:system_info}}. Figure \ref{fig:overview} illustrates the data generation process.

\subsection{Medical knowledge base}\label{ss:kb}
The dataset we propose relies on a proprietary medical knowledge base (KB) which was constructed by compiling over 20,000 medical papers including epidemiological studies, disease specific articles and meta-analysis papers. From those, the diseases' incidence and prevalence, and symptoms and risk factors' likelihood ratios across various geographies, age and sex groups were extracted. The KB was organized by diseases and reviewed by several doctors, ensuring the diseases' descriptions included atypical presentations. This KB was used to design \DXA, a rule-based AD system that has been deployed in a commercial telemedicine platform. In total, the knowledge base covers a set of 440 pathologies and 802 symptoms and antecedents. The pathologies are grouped in overlapping subgroups based on common characteristics referred to as \textit{chief complaints} \citep{Aronsky01CC, Thompson06CC}. In this work, we focus on pathologies belonging to the chief complaint related to cough, sore throat, or breathing issues. This group contains 49 pathologies covering 110 symptoms and 113 antecedents. Extending the dataset to all pathologies is left for future work.

Each disease $d$ in the KB is characterized by either an incidence rate, a prevalence rate, or both values. Both rates are conditioned on the age, sex, and geographical region of the patient. The incidence rate measures the proportion of new occurrences of a disease in a population over a period of time while the prevalence rate captures the proportion of individuals in a population that have a disease at a particular time \citep{friis2020epidemiology}. For each disease, a set of symptoms and antecedents describing the disease are provided together with their related probabilities. These probabilities are conditioned on the age and sex of the patient. Thus, the probability values $p(s | d, age, sex)$ and $p(a | d, age, sex)$ are provided for each symptom $s$ and each antecedent $a$. In some cases, a symptom $f_s$ (e.g., \textit{pain location}) may be dependent on another symptom $s$ (e.g., presence of \textit{pain}), in which case the KB provides means to extract the corresponding conditional probability $p(f_s|s, d, age, sex)$. Unlike existing datasets mentioned above, evidences (i.e., symptoms and antecedents) within this KB are of several types. They can be binary (e.g., \textit{Are you coughing?}), categorical (e.g., \textit{What is the pain intensity on a scale of 0 to 10?}), or multi-choice (e.g., \textit{Where is your pain located?}). Finally, each disease is characterized by its level of severity ranging from 1 to 5 with the lowest values describing the most severe pathologies.

\subsection{Patient generation}\label{ss:patient-generation}

A synthetic patient is an entity characterized by an \textit{age}, a \textit{sex}, a geographical region \textit{geo}, and who is suffering from a pathology $d$ and is experiencing a set $E$ of symptoms and antecedents. As a first step for modeling patients, we consider the following rule:
\begin{equation}
p(age, sex, geo, d, E) = p(age, sex, geo) \times p(d | age, sex, geo) \times p(E | d, age, sex).
\end{equation}
This formulation relies on the fact that the set of evidence $E$ experienced by a patient given a disease doesn't depend on the geographical region. In what follows, we present other rules and assumptions required to exploit the KB.

\paragraph{Assumptions on the socio-demographic data}

As we only have access to the marginal distributions, we assume that age, sex, and geographical region are independent. That is, 
\begin{equation}
p(age, sex, geo) = p(age) \times p(sex) \times p(geo).
\end{equation} 
This assumption was reviewed by doctors, who deemed it reasonable, as the diseases in the KB, unless specified, are well distributed across populations. 
The distributions of age and sex can be obtained from census data.  For this dataset, we used the 2010-2015 US Census data from the State of New York \citep{USCensus1015}. For more details, see Appendix~\ref{appx:census}. As a result of this choice, the default geographical region of a patient is ``North America''. Given that some pathologies in the KB can be contracted only if the patient is from a different geographical region, we embed the notion of recent travel when synthesizing a patient. Each synthesized patient is generated by simulating the fact that they recently travelled or not, and if they travelled, in which geographical region. We thus assume the availability of a prior distribution $p(travel)$ representing the proportion of the population travelling each month and we consider that the distribution regarding the geographical regions of destination is uniform\footnote{The dataset can be improved by using real travel and geographical destination statistics.}. Based on these assumptions, we derive the following prior distribution $p(geo)$ over the geographical regions:
\begin{itemize}
    \item Sample $u \sim \mathcal{U}(0,1)$. 
    \item If $u < p(travel)$, then randomly select a geographical region from the available set of geographical regions (see Appendix~\ref{appx:georeg}). We used $p(travel) = 0.25$ for this dataset.
    \item If $u \geq p(travel)$, then set the geographical region to be ``North America''.
\end{itemize}

\paragraph{Assumptions on pathologies}

We use the disease's incidence rate, when available, as the disease prior distribution $p(d | age, sex, geo)$, and fall back on the disease's prevalence rate when the incidence rate is missing. This is one of the major limitations of the data generation process which needs to be addressed in future work. The incidence rate can be approximated by dividing the prevalence rate with a constant factor representing the average duration of the disease, and which may be different for each disease. Out of the 49 considered diseases, 8 are affected by this limitation (see Appendix~\ref{appx:prev}).

When the resulting rate is greater than 100\% (e.g., an incidence rate of 200\% means that an individual will likely develop the pathology twice a year on average), we cap it at 100\% to avoid having a highly imbalanced dataset. Indeed, without this capping, the dataset would have been dominated by only a few pathologies, with more than half of the patients suffering from the three pathologies whose incidence rate is greater than 100\% (i.e., URTI, Viral pharyngitis, and Anemia).

Finally, the KB also contains some diseases that have extremely low incidence rates, and therefore patients suffering from those pathologies are barely generated. To increase the chance of those pathologies to be represented within the dataset, we decide to cap the rate at a minimum of 10\%. Thus, the rates used to generate the dataset lie between 10\% and 100\%. This alteration of the original rates leads to a more balanced dataset (see Figure~\ref{fig:pathoGlobalStats}).

\paragraph{Assumptions on symptoms and antecedents}

Given a sampled disease $d$, the next step is to generate all the evidences (i.e., symptoms and antecedents) the synthesized patient will be experiencing. Given that the KB doesn't contain the joint distribution of all symptoms and antecedents conditioned on the disease, sex and age, a simplifying assumption is made according to which the evidences are independent of each other when conditioned on the disease, age and sex unless explicitly stated otherwise in the KB. In other words, we have:
\begin{equation}
p(E | d, age, sex) = \prod\limits_{e \in E \backslash E_h} p(e | d, age, sex) \prod\limits_{f_s \in E_h} p(f_s | s, d, age, sex),
\end{equation} 
where $s$ is the symptom which $f_s$ depends on, $E$ is the set of evidences experienced by the patient, and $E_h$ is the subset of symptoms experienced by the patient which are dependent on other symptoms also experienced by the patient.

Some evidences, such as pain intensity, are described as integer values on a scale from 0 to 10. However, the knowledge base only provides the average value of each such evidence given the disease, the age, and the sex of the patient. To inject some randomness in the patient generation process, the values of those evidences are uniformly sampled from the interval $[\max(0, v - 3), \min(10, v + 3)]$ where v is the average value present in the knowledge base. 

Finally, in order to reflect the deployed AD system that is based on the KB, we limit to 5 the maximum number of choices associated with multi-choice evidences such as \textit{pain location}.

\paragraph{Tools}
As mentioned above, we use \Synthea to synthesize patients. To generate the value associated with a categorical or multi-choice evidence, \Synthea must be provided with a list of possible values together with their related conditional probabilities. \Synthea then goes through that list incrementally and decides, for each possible value, if it is ``on'' or not based on its probability. The process stops as soon as 1 (resp. 5) possible value(s) is (are) ``on'' for categorical (resp. multi-choice) evidences or the provided list is fully processed. In this work, the possible values of an evidence are ordered in ascending order based on their conditional probability of occurrence to make sure rare values appear in the dataset.

\subsection{Differential diagnosis generation}\label{ss:diff-generation}

As mentioned above, the KB was used to build a rule-based AD system which was deployed in a real-world telemedicine platform and which is capable of generating differential diagnoses. The system was tested in a real world clinical environment, and was vetted by doctors. In the production environment, this platform expects to be given the age, sex and an initial symptom. Based on this information, it determines a set of chief complaints (and their associated pathologies) that are compatible with the provided information, and it engages in a question-answering session with the patient to collect information about the patient's symptoms and antecedents. It then generates a differential diagnosis of ranked pathologies. We leverage this platform to compute the differential diagnosis of each synthesized patient. In order to bypass the limitations of the platform and the errors it might make in the question-answering session by omitting to ask some relevant questions, we provide all evidences in one shot, as opposed to providing each evidence iteratively upon request of the platform. More specifically, we proceed according to the following high-level steps:
\begin{itemize}
    \item We provide the age and the sex of the patient, the appropriate chief complaint, and we answer "yes" to the question "Are you consulting for a new problem?".
    \item We provide all the symptoms and antecedents experienced by the patient in one shot, at the beginning of the interaction. The motivation behind this is to make sure that the system is aware of all this information and doesn't miss on any of the patient's symptoms and antecedents.
    \item The platform may still inquire about additional questions. If that is the case, we answer "no" to those questions until we see a ``QUIT'' response from the platform or the maximum interaction length (30) is reached.
    \item When the maximum interaction length is reached, the platform does not produce a differential diagnosis. The corresponding patient is discarded from the dataset.
    \item When a "QUIT" response is provided by the platform, it contains a differential diagnosis. We further proceed by verifying if the underlying synthesized disease is part of the generated differential diagnosis. If it is not (because the platform itself is not a perfect system or because the patient didn't have enough evidences for the rule-based system to include the simulated disease in the differential diagnosis), the patient is discarded from the dataset. Each pathology within the generated differential diagnosis has a score. Those scores are normalized to obtain a probability distribution.
\end{itemize}

Given that the platform treats the provided chief complaint as a clue instead of a hard constraint and given that the platform was built to consider all 440 pathologies found in the KB, it sometimes returns a differential diagnosis that contains pathologies which do not belong to the specified chief complaint. There are several options for handling this situation: (1) create an ``other pathologies'' category and assign it the cumulative mass of the corresponding pathologies, or (2) manually remove those pathologies from the differential diagnosis and re-normalize the distribution. We opt for the second option because we want to restrict the set of pathologies to the universe of 49 pathologies used to synthesize patients. On average, we removed 1.78 ($\pm 1.68)$ pathologies from the generated differential diagnosis for an average cumulative probability mass of 0.10 ($\pm 0.11$). Statistics regarding the rank from which those pathologies are excluded are described in Appendix~\ref{appx:diffpost}.

\subsection{Dataset characteristics}

With the above assumptions and limitations, we generate, under the CC-BY licence, a dataset of roughly 1.3 million patients, where each patient is characterized by their age, sex, geographical region or recent travel history, pathology, symptoms, antecedents, as well as the related differential diagnosis. We divide the dataset into training, validation, and test subsets based on an 80\%-10\%-10\% split, using stratified sampling on the simulated pathology.

Compared to existing datasets from the AD and ASD literature, our dataset has several advantages:
\begin{itemize}
    \item To the best of our knowledge, this is the first large-scale dataset containing differential diagnoses. This information is important because doctors reason in terms of a differential and not a single pathology, as the evidence collected from a patient can sometimes point to multiple pathologies, some of which requiring additional medical exams before they can be safely ruled out.
    \item Unlike the \Symcat \citep{Symcat} and the Muzhi \citep{wei2018task} datasets, which only contain binary evidences, our dataset also includes categorical and multi-choice evidences which can naturally match the kind of questions a doctor would ask a patient, and which can lead to more efficient evidence collection. Moreover, some related evidences are defined according to a hierarchy which can be used to design systems able to interact with patients in a logical way.
    \item Our dataset makes a clear distinction between antecedents and symptoms. 
    \item Each pathology in our dataset is characterized by a severity level. This information can be used to design solutions that properly handle severe pathologies.
\end{itemize}

\subsection{Data analysis}
\label{sec:datanalysis}
We present summary statistics of the generated dataset. Those statistics are based on the entire set. Statistics on the train, validation, and test subsets are presented in Appendix~\ref{appx:subsetAnalysis}.

\textbf{Types of evidences}: The distribution of the types of evidences related to the 49 pathologies selected from the KB is provided in Table~\ref{tbl:DXAEvidendeStats}. 6.7\% of evidences are categorical and multi-choice. 

\begin{table}[htb!]
\centering
\caption{Distribution of the evidence types corresponding to the 49 pathologies selected from the KB.}
\label{tbl:DXAEvidendeStats}
\resizebox{0.6\textwidth}{!}{%
\begin{tabular}{ccccc}
\cline{2-5}
\multicolumn{1}{l}{} & \multicolumn{1}{l}{Binary} & \multicolumn{1}{l}{Categorical} & \multicolumn{1}{l}{Multi-choice} & \multicolumn{1}{l}{Total} \\ \hline
\textbf{Evidences} & 208 & 10 & 5 & 223 \\ \hline
\textbf{Symptoms} & 96 & 9 & 5 & 110 \\ \hline
\textbf{Antecedents} & 112 & 1 & 0 & 113 \\ \hline
\end{tabular}%
}
\end{table}

\textbf{Number of evidences}: Table~\ref{tbl:DXAGlobalSynPatients} shows an overview of the synthesized patients in terms of the number of simulated evidences. On average, a patient has roughly 10 symptoms and 3 antecedents.

\begin{table}[htb!]
\centering
\caption{Statistics describing the number of evidences of the synthesized patients.}
\label{tbl:DXAGlobalSynPatients}
\resizebox{0.8\textwidth}{!}{%
\begin{tabular}{cccccccc}
\cline{2-8}
\multicolumn{1}{l}{} & \multicolumn{1}{l}{Avg} & \multicolumn{1}{l}{Std dev} & \multicolumn{1}{l}{Min} & \multicolumn{1}{l}{1st quartile} & \multicolumn{1}{l}{Median} & \multicolumn{1}{l}{3rd quartile} & \multicolumn{1}{l}{Max} \\ \hline
\textbf{Evidences} & 13.56 & 5.06 & 1 & 10 & 13 & 17 & 36 \\ \hline
\textbf{Symptoms} & 10.07 & 4.69 & 1 & 8 & 10 & 12 & 25 \\ \hline
\textbf{Antecedents} & 3.49 & 2.23 & 0 & 2 & 3 & 5 & 12 \\ \hline
\end{tabular}%
}
\end{table}

\begin{figure}[htb!]
\centering
\includegraphics[width=\linewidth]{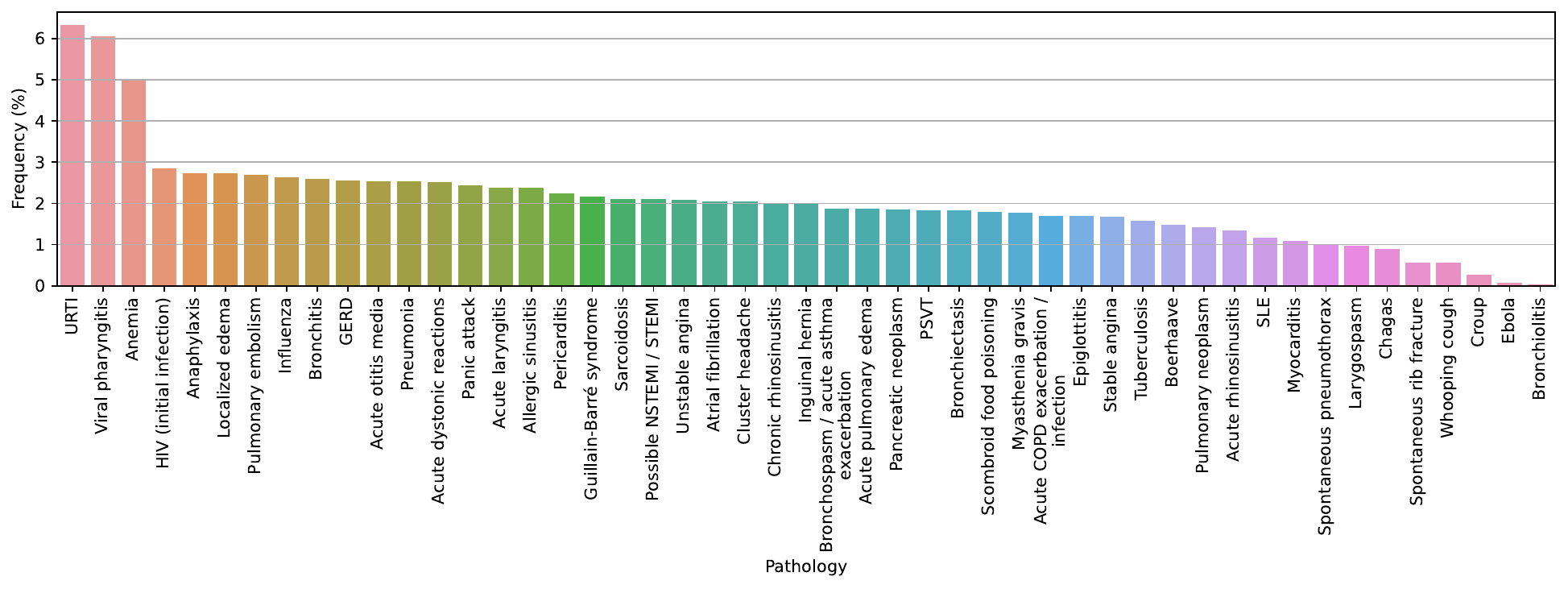}
\caption{Histogram of the pathologies experienced by the synthesized patients.}
\label{fig:pathoGlobalStats}
\end{figure}

\textbf{Pathology statistics}: Figure~\ref{fig:pathoGlobalStats} shows the histogram of the pathologies of patients in the generated dataset. Although there are three dominating pathologies (URTI, Viral pharyngitis, and Anemia), other pathologies are also  well represented.

\begin{figure}[htb!]
\centering
\includegraphics[width=0.7\textwidth]{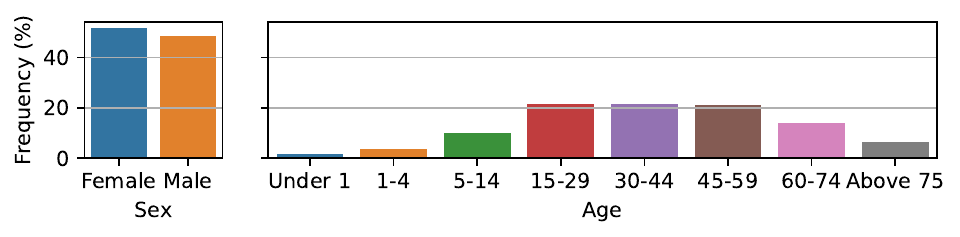}
\caption{The socio-demographic statistics of the synthesized patients.}
\label{fig:demoGlobalStats}
\end{figure}

\textbf{Socio-demographic statistics}: The statistics of the socio-demographic data of the synthesized patients are shown in Figure~\ref{fig:demoGlobalStats}. As expected, these statistics are compliant with the 2015 US Census data of the state of New York used during the generation process (see Appendix \ref{appx:census}).

\begin{figure}[htb!]
\centering
\includegraphics[width=0.7\textwidth]{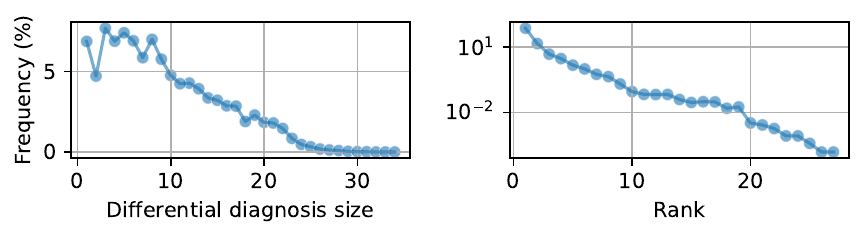}
\caption{Statistics regarding (left) the length of the differential diagnosis, and (right) the rank of the patient's simulated pathology within the differential diagnosis (y-axis on log scale).}
\label{fig:a3}
\end{figure}

\textbf{Differential diagnosis statistics}: The distribution of the length of the differential diagnosis of the synthesized patients is depicted in Figure~\ref{fig:a3} (left). As observed, the generated differential diagnosis can contain more than one pathology. It is also interesting to observe that the simulated pathology is ranked first for more than 70\% of patients (see Figure~\ref{fig:a3} (right)). 

\textbf{Sample patient}: We present a DDXPlus patient along with a doctor's analysis of this patient and the differential diagnosis. Additional samples are provided in Appendix~\ref{app:patient_samples}.

\begin{lstlisting}
Sex: F, Age: 79
Geographical region: North America
Ground truth pathology: Spontaneous pneumothorax
Symptoms:
---------
	 - I have chest pain even at rest.
	 - I feel pain.
	 - The pain is:
		 * a knife stroke
	 - The pain locations are:
		 * upper chest
		 * breast(R)
		 * breast(L)
	 - On a scale of 0-10, the pain intensity is 7
	 - On a scale of 0-10, the pain's location precision is 4
	 - On a scale of 0-10, the pace at which the pain appeared is 9
	 - I have symptoms that increase with physical exertion but alleviate with rest.
Antecedents:
-----------
	 - I have had a spontaneous pneumothorax.
	 - I smoke cigarettes.
	 - I have a chronic obstructive pulmonary disease.
	 - Some family members have had a pneumothorax.
Differential diagnosis:
----------------------
Unstable angina: 0.262, Stable angina: 0.201, Possible NSTEMI / STEMI: 0.160, GERD: 0.145, Pericarditis: 0.091, Atrial fibrillation: 0.082, Spontaneous pneumothorax: 0.060
\end{lstlisting}

The diagnosis is a spontaneous pneumothorax, which is atypical for this patient’s demography. It is rare, but possible. If we were only to collect the positive evidence and score solely on the true or false label, a doctor would find the history missing key questions for a 79 year old female presenting with chest pain that increases with exertion. We definitely need to cover the cardiovascular review more thoroughly and explore the most specific symptoms and risk factors. Optimizing for an accurate differential forces a model to ask questions that in that case would be answered “no”, but would have a really positive impact on the confidence of doctors and medical staff in the model's capability to adequately collect a medical history. Without a differential, much would be left out of the medical history.

\subsection{Dataset usage}
\label{subsec:disclamer}
We provide a description of the released dataset in Appendix~\ref{appx:release}. Based on the previous sections, it should be clear that the released dataset is meant for research purposes. Any model trained on this dataset should not be directly used in a real-world system prior to performing rigorous evaluations of the model performance and verifying that the system has proper coverage and representation of the population that it will interact with.

\section{Experiments}
\label{sec:xps}

\subsection{Models}
We consider two existing AD/ASD systems which were originally designed to predict the pathology a patient is suffering from, and adapt those systems to instead predict the differential diagnosis.
\paragraph{AARLC} AARLC \citep{yuan2021efficient} is a model that has two branches, an evidence acquisition branch, trained using RL, whose goal is to determine the next evidence to inquire about, and a classifier branch trained in a supervised way to predict the patient's disease. An adaptive approach is used to align the tasks performed by the two branches using the entropy of the distributions predicted by the classifier branch. We use the same settings as those described in \citet{yuan2021efficient} but tune the $\nu$ and $\lambda$ parameters together with the learning rate on the validation set. More details about this approach can be found in Appendix~\ref{app:asd1}.
\paragraph{BASD} This supervised learning approach builds on top of the ASD module proposed in \citet{luo-etal-2020-knowledge} (with the exception of the knowledge graph) and adds a classifier network to predict the underlying patient diagnosis at the end of the acquisition process. The agent is made of an MLP with a number $n_h$ of hidden layers of size 2048 which is tuned together with the learning rate on the validation set. More details about this approach can be found in Appendix~\ref{app:asd2}.

To assess the impact of the differential diagnosis as a training signal, we train two versions of each approach, one focused on predicting the patient's disease and one trained to predict the differential diagnosis. After training, we use the probability distribution predicted by each version at the end of an interaction as the predicted differential diagnosis. The hyper-parameters of each version are tuned independently and the resulting optimal set is used to report model performance.

\subsection{Experimental setup}
Each patient provides their age, sex, and an initial evidence to the model at the beginning of the interaction. The model then iteratively inquires about a symptom or an antecedent, until either all the relevant symptoms and antecedents have been collected or a maximum number of turns $T$ is reached. At the end of the interaction, a differential diagnosis is predicted. We use $T = 30$ in all experiments. Training is performed using a NVIDIA V100 GPU.

\subsection{Results}
An AD system is typically tasked to collect (i) relevant evidences from a patient, (ii) make accurate predictions regarding the patient's differential, and (iii) operate in a minimum number of turns. As such, we report on the interaction length (IL), and evaluate the evidence collection by measuring the recall (PER). We do not measure the evidence precision as it is sometimes necessary to ask negative questions. Additionally, we calculate the recall (DDR), precision (DDP) and F1 score (DDF1) of the differentials. 
Finally, we report the accuracy of the inclusion of the ground truth pathology (i.e., the pathology a patient was simulated from) in the predicted differential diagnosis (GTPA@1 and GTPA). Note that GTPA@1, which measures the ground truth pathology accuracy when considering the top entry in the differential, is only relevant for models trained to predict the ground truth pathology. It is not relevant for models trained to predict the differential as the ground truth pathology is not necessarily the top entry in the differential. When predicting the differential, the metrics that matter are the DD-based metrics as well as GTPA (to make sure that the ground truth pathology is in the differential). Definitions of these metrics and other details can be found in Appendix~\ref{appx:evaluation}. To compute these metrics, we post-process both the ground truth differentials and the predicted ones to remove pathologies whose mass is less than or equal to 0.01. This threshold is selected to reduce the size of the differentials by removing highly unlikely pathologies.

Table~\ref{tab:my-table} 
shows the results obtained for the two approaches. Looking at the performance of AARLC which is an AD system, we observe a significant improvement in the differential performance (i.e., the DD-based metrics) when the model is directly trained to predict the differential. This indicates the importance of using the differential as a training signal as the posterior pathology distribution generated by a model trained to predict the ground truth pathology doesn't correspond to the desired differential. We also observe a significant improvement in the recall of the positive evidence. Interaction length increases when predicting the differential as the model collects more evidence. For the BASD model, the collection of positive evidence doesn't improve; this is expected as the disease classifier branch is only enabled at the end of the interaction. But the system's explainability in the form of the differential significantly improves given that doctors can evaluate the alignment between the collected evidence and the differential. This has the potential of increasing the trust of doctors in such systems. Looking at the GPTA-based metrics, models trained to predict the patient's disease exhibit better GTPA@1 scores than the ones trained to predict the differential diagnosis. This is expected given that the patient's ground truth pathology is not always ranked at the first position in the corresponding differential diagnosis (see Figure~\ref{fig:a3} (right)). As for the GTPA metric, all approaches do well, even those trained to predict the full differential without knowing what the ground truth pathology is.

We present in Appendix~\ref{appx:model_analysis} the sequence of question-answer pairs and the differential generated by each model for the patient introduced in Section~\ref{sec:datanalysis}. The behavior of the models is commented by a doctor, including the alignment between the collected evidence and the predicted differential. 

The results that we presented should be viewed as initial baselines. The release of DDXPlus opens the possibility for improving the two models described here as well as developing new ideas for ASD and AD systems capable of collecting useful evidence and generating differentials as doctors do.

\begin{table}[htb!]
\centering
\caption{Interaction length, ground truth pathology accuracy, evidence collection, and differential diagnosis metrics of the trained agents as measured on the test set. Diff indicates whether the agent was trained to predict the differential diagnosis ($\checkmark$) or not ($\times$). Except IL, all values are expressed in \%. Values indicate the average of 3 runs, and numbers
in brackets indicate 95\% confidence intervals.
A higher GPTA@1 is desired for models trained to predict the ground truth pathology, while a higher GTPA is desired for models trained to predict a differential. For all models, higher PER, DDR, DDP and DDF1 is better.}
\label{tab:my-table}
\vspace{1mm}
\resizebox{\textwidth}{!}{%
\begin{tabular}{c|c|ccccccc}
\hline
\textbf{Method} & \textbf{Diff} & \textbf{IL} & \textbf{GTPA@1} & \textbf{GTPA} & \textbf{PER}& \textbf{DDR} & \textbf{DDP} & \textbf{DDF1} \\ \hline
        \multirow{2}{*}{AARLC} & $\checkmark$ & 25.75 (2.75) & 75.39 (5.53) & 99.92 (0.03) &  54.55 (14.73) & \textbf{97.73 (1.21)} & 69.53 (8.51) & \textbf{78.24 (6.82)} \\ \cline{2-9}
         & $\times$ & \textbf{6.73 (1.35)} & {99.21} (0.78) & \textbf{99.97 (0.01)} & 32.78 (13.92) & 21.96 (0.30) & \textbf{99.19 (0.56)} & 31.28 (0.38)\\ \hline
         \hline
        \multirow{2}{*}{BASD}  & $\checkmark$ & {17.86 (0.88)} & 67.71 (1.19) & {99.30} (0.27) & 88.18 (1.12) & \textbf{85.03 (3.46)} & 88.34 (1.14) & \textbf{83.69 (1.57)}  \\ \cline{2-9}
        & $\times$ & 17.99 (3.57) & {97.15} (1.70) & {98.82} (1.03) & {88.45} (5.78) & 21.89 (0.19) & \textbf{99.38 (0.07)} & 31.31 (0.29) \\ \hline
\end{tabular}%
}
\end{table}

\section{Conclusion}
\label{sec:conclusion}
We release a large-scale benchmark dataset of roughly 1.3 million patients suffering from pathologies that include cough, sore throat or breathing problems as symptoms. The dataset contains binary, categorical and multi-choice symptoms and antecedents. Each patient within the dataset is characterized by their age, sex, geographical region or recent travel history, pathology, symptoms, antecedents, as well as the related differential diagnosis. We extended two existing approaches from the AD/ASD literature (based on RL and non-RL settings) to leverage the differential diagnosis available in the proposed dataset. The obtained results provide empirical evidence that using the differential diagnosis as a training signal enhances the performance of such systems or helps doctors better understand the reasoning of those systems by evaluating the collected evidence, the predicted differential and their alignment.
We hope that this dataset will encourage the research community to develop automatic diagnosis systems that can be trusted by medical practitioners as the latter operate with differential diagnoses when interacting with real patients. In constructing this dataset, we guarantee that the differentials are informed by all relevant positive evidences the patients may have. However, while the rule-based AD system which generates the differentials may further inquire about negative evidences, it is not guaranteed that all relevant negative evidences, i.e., evidences that may change the differentials, are considered. Doing so would be much harder practically as the scope of negative evidences is much larger, and we leave this to future work. While extending the above mentioned approaches, we considered all pathologies as equally important. But in general, when establishing a differential diagnosis, medical practitioners likely ask questions to specifically explore and rule out severe pathologies. The proposed dataset has a severity flag associated with each pathology. This leaves room for exploring approaches that better handle severe pathologies. Finally, we would like to emphasize that this dataset should not be used to train and deploy a model prior to performing rigorous evaluations of the model performance and verifying that the system has proper coverage and representation of the population that it will interact with.

\begin{ack}
We would like to thank Dialogue Health Technologies Inc. for providing us access to the knowledge base and the rule-based AD system used throughout this work. We would also like to thank Quebec's Ministry of Economy and Innovation and Invest AI for their financial support.
\end{ack}

\clearpage

\bibliographystyle{plainnat}
\bibliography{references}

\section*{Checklist}

\begin{enumerate}

\item For all authors...
\begin{enumerate}
  \item Do the main claims made in the abstract and introduction accurately reflect the paper's contributions and scope?
    \answerYes{See Sections~\ref{sec:ds} and~\ref{sec:xps}.}
  \item Did you describe the limitations of your work?
    \answerYes{See Sections~\ref{sec:ds} and~\ref{sec:conclusion}.}
  \item Did you discuss any potential negative societal impacts of your work?
    \answerYes{See Sections~\ref{subsec:disclamer} and~\ref{sec:conclusion}.}
  \item Have you read the ethics review guidelines and ensured that your paper conforms to them?
    \answerYes{See Sections~\ref{sec:ds} and~\ref{sec:conclusion}.}
\end{enumerate}

\item If you are including theoretical results...
\begin{enumerate}
  \item Did you state the full set of assumptions of all theoretical results?
    \answerNA{}
	\item Did you include complete proofs of all theoretical results?
    \answerNA{}
\end{enumerate}

\item If you ran experiments (e.g. for benchmarks)...
\begin{enumerate}
  \item Did you include the code, data, and instructions needed to reproduce the main experimental results (either in the supplemental material or as a URL)?
    \answerYes{See the abstract for a link to the dataset and the code is present in the supplementary. The code contains a readme with instructions needed to reproduce the main experimental results.}
  \item Did you specify all the training details (e.g., data splits, hyperparameters, how they were chosen)?
    \answerYes{See Sections~\ref{sec:ds} and~\ref{sec:xps}.}
	\item Did you report error bars (e.g., with respect to the random seed after running experiments multiple times)?
    \answerYes{See Table~\ref{tab:my-table} in Section~\ref{sec:xps}.}
	\item Did you include the total amount of compute and the type of resources used (e.g., type of GPUs, internal cluster, or cloud provider)?
    \answerYes{See Section~\ref{sec:xps}.}
\end{enumerate}

\item If you are using existing assets (e.g., code, data, models) or curating/releasing new assets...
\begin{enumerate}
  \item If your work uses existing assets, did you cite the creators?
    \answerYes{See Section~\ref{sec:ds} for \Synthea and Section~\ref{sec:xps} for code related citations.}
  \item Did you mention the license of the assets?
    \answerYes{See Section~\ref{sec:ds} for the dataset and its licensing. The license for the code is attached with the code in the supplementary.}
  \item Did you include any new assets either in the supplemental material or as a URL?
    \answerYes{We curated a new dataset whose link is in the abstract. We also include code in the supplementary.}
  \item Did you discuss whether and how consent was obtained from people whose data you're using/curating?
    \answerNA{We curated a synthetic dataset.}
  \item Did you discuss whether the data you are using/curating contains personally identifiable information or offensive content?
    \answerNA{We used a synthetic dataset.}
\end{enumerate}

\item If you used crowdsourcing or conducted research with human subjects...
\begin{enumerate}
  \item Did you include the full text of instructions given to participants and screenshots, if applicable?
    \answerNA{}
  \item Did you describe any potential participant risks, with links to Institutional Review Board (IRB) approvals, if applicable?
    \answerNA{}
  \item Did you include the estimated hourly wage paid to participants and the total amount spent on participant compensation?
    \answerNA{}
\end{enumerate}

\end{enumerate}

\clearpage
\appendix
\normalsize

\section{Additional information on the KB and rule-based AD system used to generate DDXPlus}
\label{appx:system_info}
Doctors reviewed over 2 years of relevant papers on the diseases used to create the KB. The papers along with the medical experts' knowledge and experience were used to extract the typical and atypical presentations of diseases, along with the relevant symptom and antecedent distributions to build accurate disease models based on geography, demography and baseline patient characteristics. The process was exhaustive and independently validated by the doctors, where agreement was sought for the presentation of every disease in the database. This process therefore uses the clinician experience to ensure that the diseases are accurately depicted across their usual and unusual presentations. 

Some of the symptoms and antecedents in the KB were defined to be categorical or multi-choice to ensure efficient and proper coverage of important evidence. Since almost 60\% of all presentations in real life will include pain, a great level of attention was given to the description of pain for each disease where this symptom can be found \citep{ cordell2002high, mura2017prospective}. This feature includes the localization, radiation, intensity, subjective characterization of common pain description by patients, a precision feature (very small area to diffuse) and a rapidity of onset. The pain symptom encompasses all these sub-features and was created using domain experts and medical journal articles looking at disease presentations. This would be extremely hard to derive from another dataset, since we have not seen other ones cover the pain description extensively, although one of the most important symptoms a clinician will spend a good amount of time defining clearly with the patient.
A similar process of specification was used for skin rashes, with the rash description including the usual dermatological lesion characterizations used by clinicians when evaluating a patient.

The rule-based AD system is a statistical engine that uses the patient's response to generate a differential diagnosis in real-time. The engine has phases, where the first phase seeks to ask questions that have the highest probability of ruling out the most diseases in the initial differential. Subsequently, the engine seeks answers to questions linked to diseases in the differential that represent the highest risk in terms of mortality and morbidity. Finally, the engine seeks to ask specific questions about the personal risk factors and antecedents for the top 5 diseases in the estimated differential.

The engine was built for primary care and acute care settings, having in mind the goal of gathering a medical history that is as close as possible to the one clinicians would gather when evaluating a patient. 

The engine was tested on real patients in an acute primary care setting and the collected history was evaluated by doctors on a scale of pertinence and completeness.
The differential of the engine was compared to the clinician’s differential, who evaluated the patient with the usual clinical flow, blinded to the evidence collected by the engine initially.

\section{Demographic statistics from census data}
\label{appx:census}
To synthesize patients, one needs to have access to the prior distributions of the age and sex of a population of interest. In this work, we rely on the 2010-2015 US census data \cite{USCensus1015} for the state of New York. Table~\ref{tbl:census} describes the corresponding statistics.

\begin{table}[htb!]
\centering
\caption{The 2010-2015 census data of the state of New York.}
\label{tbl:census}
\begin{tabular}{ccc}
\cline{2-3}
 & \textbf{Category} & \textbf{Frequency} \\ \hline
\multirow{2}{*}{\textbf{Sex}} & Male & 0.4836 \\ \cline{2-3} 
 & Female & 0.5164 \\ \hline
\multirow{8}{*}{\textbf{Age}} & Less than 1-year & 0.0154 \\ \cline{2-3} 
 & 1-4-years & 0.0461 \\ \cline{2-3} 
 & 5-14-years & 0.1146 \\ \cline{2-3} 
 & 15-29-years & 0.2132 \\ \cline{2-3} 
 & 30-44-years & 0.2025 \\ \cline{2-3} 
 & 45-59-years & 0.2042 \\ \cline{2-3} 
 & 60-74-years & 0.1399 \\ \cline{2-3} 
 & 75-years and more & 0.0641 \\ \hline
\end{tabular}
\end{table}

\section{Geographical regions}
\label{appx:georeg}
The following geographical regions are covered in the dataset: North Africa, West Africa, South Africa, Central America, North America, South America, Asia, South East Asia, The Caribbean, Europe, Oceania.

\section{Missing incidence rates}
\label{appx:prev}
The 8 pathologies whose incidence rates were not present for all the combinations of age, sex, and geographical regions are: Anemia, Inguinal hernia, Anaphylaxis, Allergic sinusitis, Chagas, Tuberculosis, Ebola, Chronic rhinosinusitis. Some are very rare infectious diseases that follow an epidemic pattern which varies over time periods, making it hard to track incidence. Others are very mild diseases and are not tracked as they represent no public health interest.

\section{Differential diagnosis post-processing}
\label{appx:diffpost}
As mentioned previously, pathologies that are part of the differential diagnosis generated by the rule-based AD platform and that do not belong to the set of the 49 considered pathologies are excluded from the differential diagnosis. Table~\ref{tbl:exc_rank_diff} describes, for each rank in the differential diagnosis, the proportion of patients for which the pathology at that rank is excluded from the differential diagnosis.

\begin{table}[htb!]
\centering
\caption{Proportion (\%) of the patients for which pathologies are excluded at each rank from the differential diagnosis returned by the rule-based AD platform.}
\small
\begin{tabular}{cc}
\hline
\multicolumn{1}{c}{\textbf{Rank}} & \textbf{Proportion} \\ \hline
1                                   & 1.09       \\ \hline
2                                   & 11.23       \\ \hline
3                                   & 15.73       \\ \hline
4                                   & 13.78     \\ \hline
5                                   & 14.53      \\ \hline
6                                   & 11.92     \\ \hline
7                                   & 13.12     \\ \hline
8                                   & 10.47      \\ \hline
9                                   & 9.23     \\ \hline
10                                   & 8.16     \\ \hline
11                                  & 9.47    \\ \hline
12                                  & 7.00     \\ \hline
13                                  & 5.92       \\ \hline
14                                  & 6.11     \\ \hline
15                                  & 5.49     \\ \hline
16                                  & 5.86    \\ \hline
17                                  & 5.42     \\ \hline
18                                  & 4.52     \\ \hline
19                                  & 3.67     \\ \hline
20                                  & 3.37     \\ \hline
21                                  & 3.07     \\ \hline
22                                  & 2.37    \\ \hline
23                                  & 2.14     \\ \hline
24                                  & 1.70    \\ \hline
25                                  & 1.17    \\ \hline
26                                  & 0.71    \\ \hline
27                                  & 0.42    \\ \hline
28                                  & 0.24    \\ \hline
29                                  & 0.13   \\ \hline
30                                  & 0.07   \\ \hline
31                                  & 0.04    \\ \hline
32                                  & 0.03   \\ \hline
33                                  & 0.02  \\ \hline
34                                  & 0.01  \\ \hline
35                                  & 4.49e-03  \\ \hline
36                                  & 1.76e-03  \\ \hline
37                                  & 3.90e-04   \\ \hline
\end{tabular}
\label{tbl:exc_rank_diff}
\end{table}

\section{Subset data analysis}
\label{appx:subsetAnalysis}

\subsection{Evidence statistics}
\label{appx:evisubsets}

The statistics of the evidences experienced by the synthesized patients belonging to the train, validation, and test subsets are presented in Tables~\ref{tbl:DXATrainSynPatients},~\ref{tbl:DXAValSynPatients}, and~\ref{tbl:DXATestSynPatients} respectively. As illustrated, the evidences are similarly distributed across the three subsets.

\begin{table}[htb!]
\centering
\scriptsize
\caption{Statistics describing the number of evidences of the synthesized patients for the training set.}
\begin{tabular}{lccc}
\cline{2-4}
                                            & \textbf{Evidences} & \textbf{Symptoms} & \textbf{Antecedents} \\ \hline
\multicolumn{1}{l}{\textbf{Avg}}          & 13.52               & 10.03              & 3.49                \\ \hline
\multicolumn{1}{l}{\textbf{Std dev}}      & 5.08               & 4.71              & 2.23                 \\ \hline
\multicolumn{1}{l}{\textbf{Min}}          & 1                  & 1                 & 0                    \\ \hline
\multicolumn{1}{l}{\textbf{1st quartile}} & 10                  & 8                 & 2                    \\ \hline
\multicolumn{1}{l}{\textbf{Median}}       & 13                  & 10                 & 3                    \\ \hline
\multicolumn{1}{l}{\textbf{3rd quartile}} & 17                 & 12                 & 5                    \\ \hline
\multicolumn{1}{l}{\textbf{Max}}          & 36                 & 25                & 12                   \\ \hline
\end{tabular}
\label{tbl:DXATrainSynPatients}
\end{table}

\begin{table}[htb!]
\centering
\scriptsize
\caption{Statistics describing the number of evidences of the synthesized patients for the validation set.}
\begin{tabular}{lccc}
\cline{2-4}
                                            & \textbf{Evidences} & \textbf{Symptoms} & \textbf{Antecedents} \\ \hline
\multicolumn{1}{l}{\textbf{Avg}}          & 13.76               & 10.27            & 3.49                 \\ \hline
\multicolumn{1}{l}{\textbf{Std dev}}      & 5.01               & 4.61              & 2.23                 \\ \hline
\multicolumn{1}{l}{\textbf{Min}}          & 1                  & 1                 & 0                    \\ \hline
\multicolumn{1}{l}{\textbf{1st quartile}} & 10                  & 8                 & 2                    \\ \hline
\multicolumn{1}{l}{\textbf{Median}}       & 13                  & 10                 & 3                    \\ \hline
\multicolumn{1}{l}{\textbf{3rd quartile}} & 17                 & 12                 & 5                    \\ \hline
\multicolumn{1}{l}{\textbf{Max}}          & 34                 & 25                & 12                   \\ \hline
\end{tabular}
\label{tbl:DXAValSynPatients}
\end{table}

\begin{table}[htb!]
\centering
\scriptsize
\caption{Statistics describing the number of evidences of the synthesized patients for the test set.}
\begin{tabular}{lccc}
\cline{2-4}
                                            & \textbf{Evidences} & \textbf{Symptoms} & \textbf{Antecedents} \\ \hline
\multicolumn{1}{l}{\textbf{Avg}}          & 13.72               & 10.23            & 3.50                 \\ \hline
\multicolumn{1}{l}{\textbf{Std dev}}      & 5.02               & 4.63              & 2.23                 \\ \hline
\multicolumn{1}{l}{\textbf{Min}}          & 1                  & 1                 & 0                    \\ \hline
\multicolumn{1}{l}{\textbf{1st quartile}} & 10                  & 8                 & 2                    \\ \hline
\multicolumn{1}{l}{\textbf{Median}}       & 13                  & 10                 & 3                    \\ \hline
\multicolumn{1}{l}{\textbf{3rd quartile}} & 17                 & 12                 & 5                    \\ \hline
\multicolumn{1}{l}{\textbf{Max}}          & 35                 & 25                & 12                   \\ \hline
\end{tabular}
\label{tbl:DXATestSynPatients}
\end{table}

\subsection{Pathology statistics}
\label{appx:pathosubsets}

\begin{figure}[htb!]
\centering
\includegraphics[width=\linewidth]{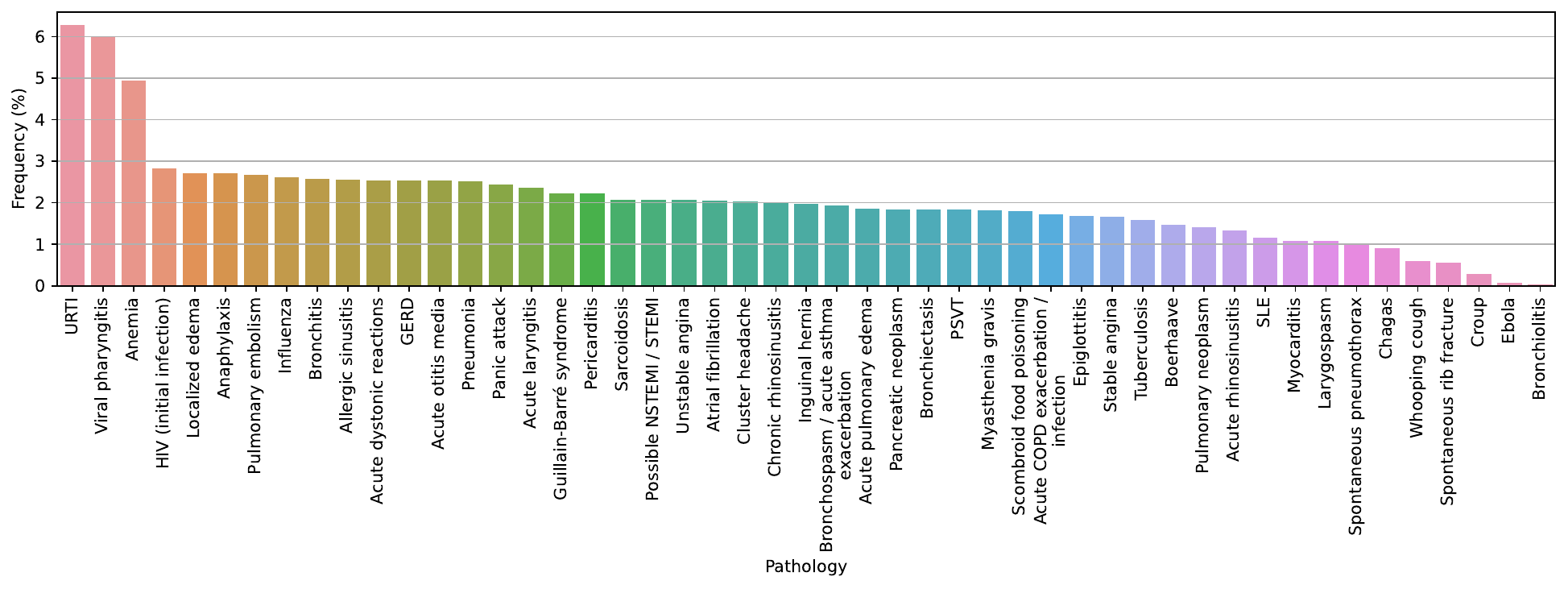}
\caption{Histogram of the patient pathologies in the training set.}
\label{fig:pathoTrainStats}
\end{figure}

\begin{figure}[htb!]
\centering
\includegraphics[width=\linewidth]{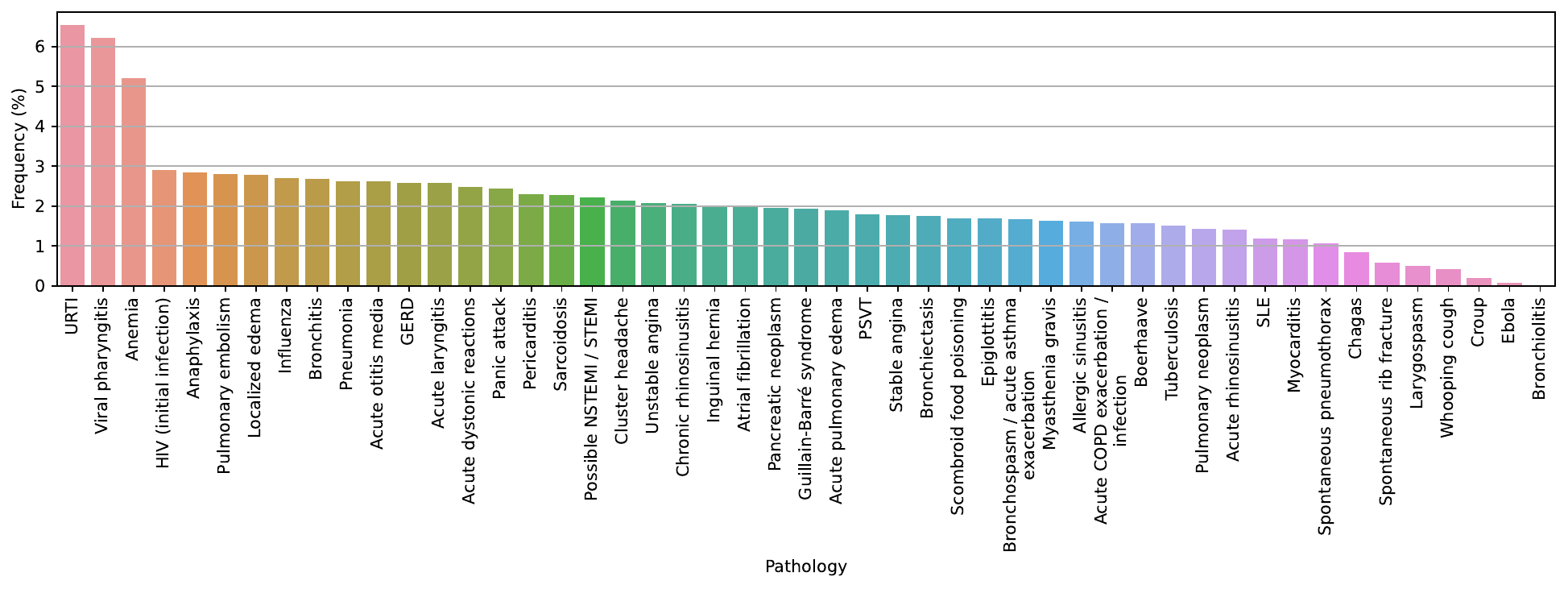}
\caption{Histogram of the patient pathologies in the validation set.}
\label{fig:pathoValidStats}
\end{figure}

\begin{figure}[htb!]
\centering
\includegraphics[width=\linewidth]{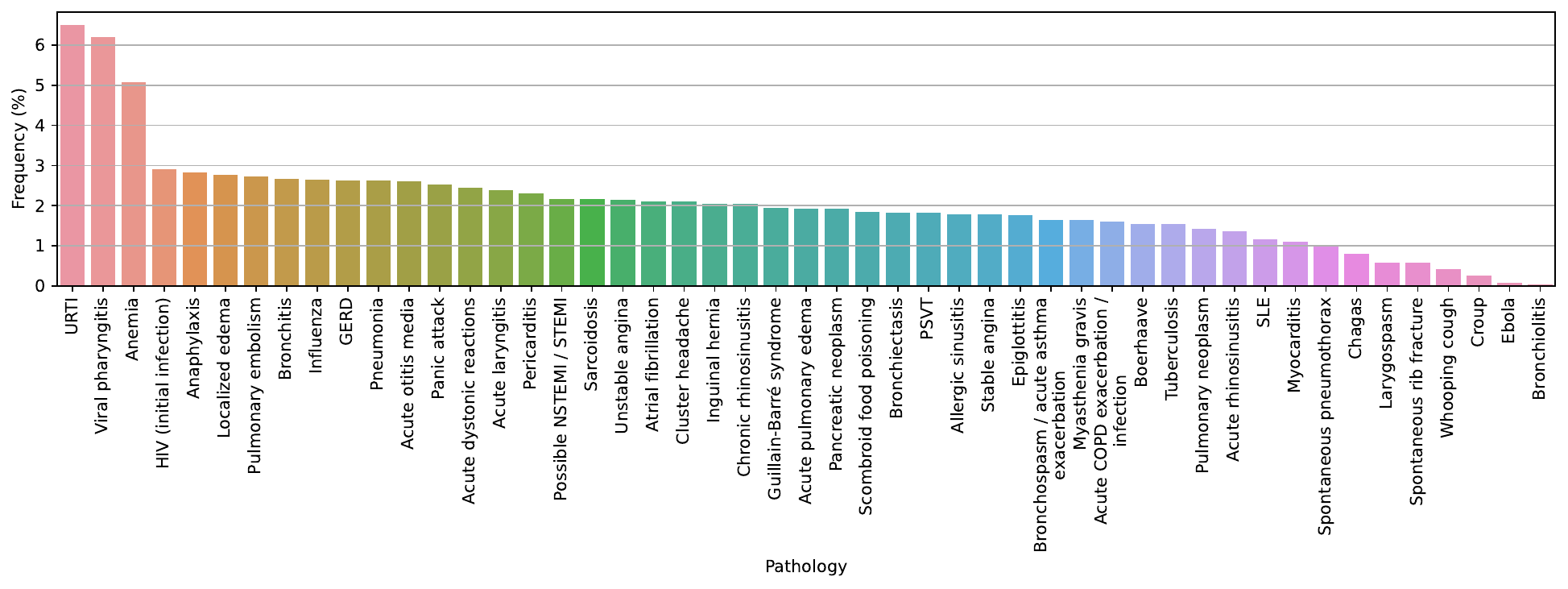}
\caption{Histogram of the patient pathologies in the test set.}
\label{fig:pathoTestStats}
\end{figure}

Figures~\ref{fig:pathoTrainStats}, ~\ref{fig:pathoValidStats}, and~\ref{fig:pathoTestStats} depict the histograms of the pathologies experienced by the synthesized patients in the train, validation, and test subsets. The pathologies are evenly distributed across the three subsets.

\subsection{Socio-demographic statistics}
\label{appx:demosubsets}

\begin{figure}[htb!]
\centering
\includegraphics[width=0.7\textwidth]{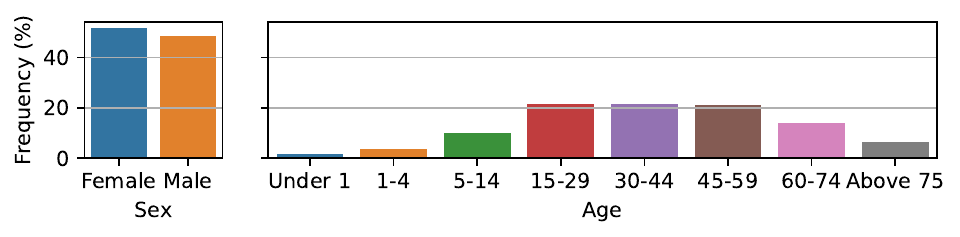}
\caption{The socio-demographic statistics of the synthesized patients from the training set.}
\label{fig:demoTrainStats}
\end{figure}

\begin{figure}[htb!]
\centering
\includegraphics[width=0.7\textwidth]{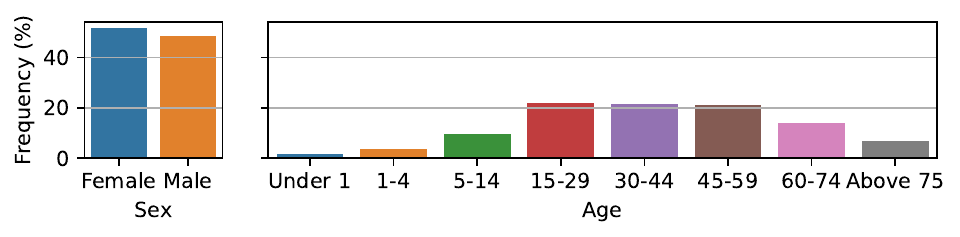}
\caption{The socio-demographic statistics of the synthesized patients from the validation set.}
\label{fig:demoValidateStats}
\end{figure}

\begin{figure}[htb!]
\centering
\includegraphics[width=0.7\textwidth]{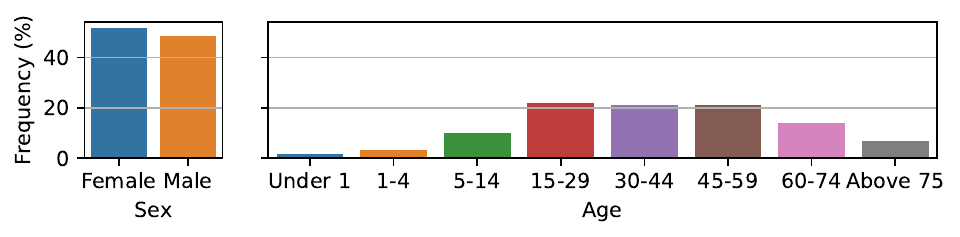}
\caption{The socio-demographic statistics of the synthesized patients from the test set.}
\label{fig:demoTestStats}
\end{figure}

The socio-demographic statistics of the synthesized patients from the train, validation and test subsets are respectively illustrated in Figures~\ref{fig:demoTrainStats}, \ref{fig:demoValidateStats}, and \ref{fig:demoTestStats}.

\subsection{Differential diagnosis statistics}
 
\begin{figure}[htb!]
\centering
\includegraphics[width=0.7\textwidth]{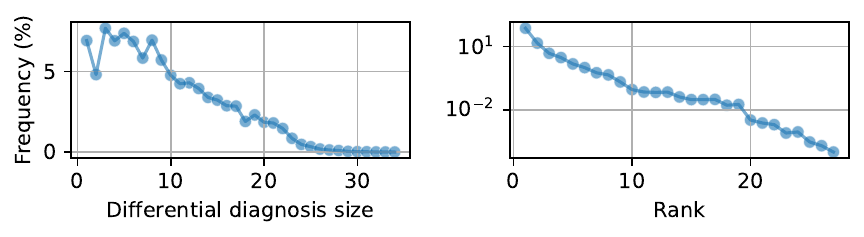}
\caption{Statistics regarding (left) the length of the differential diagnosis, and (right) the rank of the patient's simulated pathology within the differential diagnosis (y-axis on log scale), in the training set.}
\label{fig:DiffLenRankTrainStats}
\end{figure}

\begin{figure}[htb!]
\centering
\includegraphics[width=0.7\textwidth]{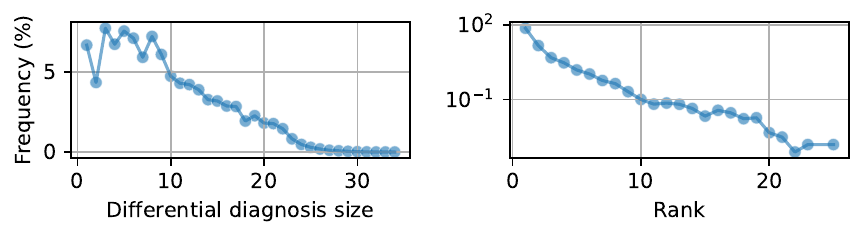}
\caption{Statistics regarding (left) the length of the differential diagnosis, and (right) the rank of the patient's simulated pathology within the differential diagnosis (y-axis on log scale), in the validation set.}
\label{fig:DiffLenRankValidateStats}
\end{figure}

\begin{figure}[htb!]
\centering
\includegraphics[width=0.7\textwidth]{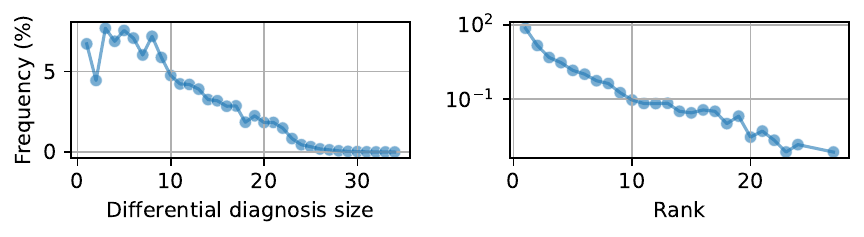}
\caption{Statistics regarding (left) the length of the differential diagnosis, and (right) the rank of the patient's simulated pathology within the differential diagnosis (y-axis on log scale), in the test set.}
\label{fig:DiffLenRankTestStats}
\end{figure}

 The histograms of the length of the differential diagnosis from the train, validation, and test subsets are respectively illustrated in Figures~\ref{fig:DiffLenRankTrainStats} (left), ~\ref{fig:DiffLenRankValidateStats} (left), and~\ref{fig:DiffLenRankTestStats} (left). Similarly, the histograms of the rank of the simulated pathology within the differential diagnosis from the train, validation, and test subsets are respectively depicted in Figures~\ref{fig:DiffLenRankTrainStats} (right), ~\ref{fig:DiffLenRankValidateStats} (right), and~\ref{fig:DiffLenRankTestStats} (right).

\section{Baselines}
\label{app:asd}
The code to reproduce the main results from Table~\ref{tab:my-table} can be found at \url{https://github.com/bruzwen/ddxplus}.

\subsection{AARLC}
\label{app:asd1}
AARLC (short for Adaptive Alignment of Reinforcement Learning and Classification) \citep{yuan2021efficient} is a model that has two branches, an evidence acquisition branch, trained using RL, whose goal is to determine the next evidence to inquire about, and a classifier branch trained in a supervised way to predict the patient's disease. An adaptive approach is used to align the tasks performed by the two branches using the entropy of the distributions predicted by the classifier branch.

To train AARLC with differentials, we make several changes, in addition to replacing the ground truth pathology with the ground truth differential probabilities as the classifier's training objective:
\begin{itemize}
    \item Instead of updating the stopping threshold $K$ when the predicted pathology matches the ground truth pathology, we update it when the set $D^i$ of diseases in the ground truth differential is identical to the set of top-$|D^i|$ predicted diseases. We make this change because AARLC designs the threshold to be updated when the predicted disease is correct, and therefore if the differential is replacing the ground truth pathology as the target, it should also replace it as the standard of correctness.
    \item Second, now that the agent does not focus on predicting one single disease, it is no longer reasonable to only update the threshold associated with one disease. Therefore, we instead use one global threshold that is not associated with any particular disease, and update it every time the aforementioned condition is met.
    \item Similar to the condition of updating the threshold, we change the condition under which a positive reward is given to the agent, as part of $r_p$, for making the correct diagnosis. We give the positive reward when the set $D^i$ of diseases in the ground truth differential is identical to the set of top-$|D^i|$ predicted diseases.
\end{itemize}

\subsection{BASD approach}
\label{app:asd2}
The BASD (short for Baseline ASD) agent consists of an MLP network with 2 prediction branches:
\begin{itemize}
    \item a policy branch whose role is to predict whether to stop or continue the interaction, and if the latter, what evidence to inquire about next;
    \item a classifier branch to make a prediction regarding the underlying patient disease.
\end{itemize}

To train the network, we simulate dialogue states together with their target values. Assuming that a given patient has $n$ evidences that they are experiencing, we simulate a dialogue state as follows:
\begin{enumerate}
    \item Randomly select $p \in [1, n]$ representing the number of positive evidences already inquired about. Sample $p$ evidences from the ones experienced by the patient and set them in the simulated dialog state.
    \item Randomly select $q \in [0, T - p)$ representing the number of negative evidences already inquired, where T is the maximum number of allowed dialog turns. Sample $q$ evidences from the ones not experienced by the patient and set them in the simulated dialog state.
    \item If $p = n$, set the target of the policy branch to "stop"; otherwise, set the target to be one of the experienced evidences that was not sampled at step 1.
    \item Set the classifier branch target to be the ground truth pathology or the ground truth differential.
\end{enumerate}

Both branches are trained using the cross-entropy loss function and  the classifier branch is only updated when the target of the policy branch is set to "stop".

\section{Evaluation metrics}\label{appx:evaluation}
This section describes the metrics used to evaluate the performance of trained agents. The differentials generated by the trained models as well as the ground truth differentials are post-processed to remove pathologies whose mass is less than or equal to 0.01. This threshold, approved by our collaborating physician, is selected to reduce the size of the differentials by removing highly unlikely pathologies. Let $D$ be the number of patients, $T^{i}$ be the set of evidences collected by an agent from the $i^{th}$ patient (including the first evidence provided by the patient), $Y^i$ be the ground truth differential, and $\hat{Y}^i$ be the pathology distribution generated by the agent for that patient.

\textbf{Interaction length (IL):} The average interaction length is defined as:
\begin{equation}
    IL =  \frac{1}{|D|} \Sigma_{i=1}^{|D|}|T^{i}|.
\end{equation}




\textbf{Differential diagnosis recall (DDR): } This metric measures the recall of the differential diagnosis predicted by the agent with respect to the ground truth differential:
\begin{equation}
    DDR = \frac{1}{|D|}\Sigma_{i=1}^{|D|}\frac{|\hat{Y}^{i} \cap Y^{i}|}{|Y^{i}|}.\label{eq:DDRecallFull}
\end{equation}

\textbf{Differential diagnosis precision (DDP):} This metric measures the precision of the differential diagnosis predicted by an agent with respect to the ground truth differential:
\begin{equation}
    DDP = \frac{1}{|D|}\Sigma_{i=1}^{|D|}\frac{|\hat{Y}^{i} \cap Y^{i}|}{|\hat{Y}^{i}|}.\label{eq:DDPFull}
\end{equation}

\textbf{Differential diagnosis F1 (DDF1):} We combine the DDR and DDP metrics to compute the F1 score of the differential diagnosis.

\textbf{Ground truth pathology accuracy (GTPA@$k$ and GTPA):} The GTPA@$k$ metric measures whether the differential diagnosis predicted by an agent contains the pathology $p$ a patient was simulated from within its top-$k$ entries ${\hat{Y}_k^{i}}$:
\begin{equation}
    GTPA@k = \frac{1}{|D|}\Sigma_{i=1}^{|D|} \mathds{1}_{\hat{Y}_k^{i}}(p^{i}),\label{eq:GTPA@k}
\end{equation}
where $\mathds{1}$ is the indicator function.

Similarly, the GTPA metric measures whether the differential diagnosis predicted by an agent contains the pathology $p$ a patient was simulated from:
\begin{equation}
    GTPA = \frac{1}{|D|}\Sigma_{i=1}^{|D|} \mathds{1}_{\hat{Y}^{i}}(p^{i}),\label{eq:GTPA}
\end{equation}
where $\mathds{1}$ is still the indicator function.

\textbf{Positive evidence recall (PER):} Let us suppose that the $i^{th}$ patient is experiencing the set $\mathcal{S}^{i}$ of symptoms and the set $\mathcal{A}^{i}$ of antecedents. Also let us assume that the agent inquires the set $\mathcal{\hat{S}}^{i}$ (resp. $\mathcal{\hat{A}}^{i}$) of symptoms (resp. antecedents) from which $\mathcal{\hat{S}}_{+}^{i} \subseteq \mathcal{S}^{i}$ (resp. $\mathcal{\hat{A}}_{+}^{i} \subseteq \mathcal{A}^{i}$) is the set of symptoms (resp. antecedents) experienced by the $i^{th}$ patient. Then, the recall for the positive evidences is calculated as:
\begin{equation}
    PER = \frac{1}{|D|} \Sigma_{i=1}^{|D|} PER_i, \qquad\text{ where  } PER_i = \frac{|\mathcal{\hat{S}}_{+}^{i} \cup \mathcal{\hat{A}}_{+}^{i} |}{|\mathcal{{S}}^{i} \cup \mathcal{{A}}^{i}|}.
\end{equation}

While collecting positive evidence is important, it is not sufficient as this evidence paints an incomplete picture to the clinical team. Clinicians want to make sure other pathologies were considered and thoroughly evaluated during their interaction with the patient. Those include severe pathologies, less prevalent pathologies, and pathologies which are similar to the one the patient is suffering from but are very different in their management. As such, clinicians inquire about other evidences and ensure they are not present. We therefore do not report the evidence precision as a metric. Defining metrics for evaluating questions about negative evidence is left for future work.



\section{Dataset release} 
\label{appx:release}
We are releasing under the CC-BY licence a new large-scale dataset for Automatic Symptom Detection (ASD) and Automatic Diagnosis (AD) systems in the medical domain. The dataset contains patients synthesized using a proprietary medical knowledge base and a commercial rule-based AD system. Patients in the dataset are characterized by their socio-demographic data, a pathology they are suffering from, a set of symptoms and antecedents related to this pathology, and a differential diagnosis. The symptoms and antecedents can be binary, categorical and multi-choice, with the potential of leading to more efficient and natural interactions between ASD/AD systems and patients.  Moreover, some symptoms are organized in a hierarchy, making it possible to design systems able to interact with patients in a logical way. Finally, each disease is characterized by its level of severity. To the best of our knowledge, this is the first large-scale dataset that includes the differential diagnosis, and non-binary symptoms and antecedents. The dataset can be downloaded at \href{https://figshare.com/articles/dataset/DDXPlus_Dataset/20043374}{https://figshare.com/articles/dataset/DDXPlus\_Dataset/20043374}.

\subsection{Dataset documentation}
 In what follows, we use the term ``evidence'' as a general term to refer to a symptom or an antecedent. The dataset contains the following files:
\begin{itemize}
    \item \texttt{release\_evidences.json}: a JSON file describing all possible evidences considered in the dataset.
    \item \texttt{release\_conditions.json}: a JSON file describing all pathologies considered in the dataset.
    \item \texttt{release\_train\_patients.zip}: a CSV file containing the patients of the training set.
    \item \texttt{release\_validate\_patients.zip}: a CSV file containing the patients of the validation set.
    \item \texttt{release\_test\_patients.zip}: a CSV file containing the patients of the test set.
\end{itemize}

\paragraph{Evidence description}
Each evidence in the \texttt{release\_evidences.json} file is described using the following entries:
\begin{itemize}
    \item \texttt{name}: name of the evidence.
    \item \texttt{code\_question}: a code allowing to identify which evidences are related. Evidences having the same \texttt{code\_question} form a group of related evidences. The value of the \texttt{code\_question} refers to the evidence that needs to be simulated/activated for the other members of the group to be eventually simulated.
    \item \texttt{question\_fr}: the query, in French, associated to the evidence.
    \item \texttt{question\_en}: the query, in English, associated to the evidence.
    \item \texttt{is\_antecedent}: a flag indicating whether the evidence is an antecedent or a symptom.
    \item \texttt{data\_type}: the type of the evidence. We use ``B'' for binary, ``C'' for categorical, and ``M'' for multi-choice.
    \item \texttt{default\_value}: the default value of the evidence. If this value is used to characterize the evidence, then it is as if the evidence was not synthesized.
    \item \texttt{possible-values}: the possible values for the evidence. Only valid for categorical and multi-choice evidences.
    \item \texttt{value\_meaning}: The meaning, in French and English, of each code that is part of the \texttt{possible-values} field. Only valid for categorical and multi-choice evidences.
\end{itemize}

\paragraph{Pathology description}
The file \texttt{release\_conditions.json} contains information about the pathologies patients in the datasets may suffer from. Each pathology has the following attributes:
\begin{itemize}
    \item \texttt{condition\_name}: name of the pathology.
    \item \texttt{cond-name-fr}: name of the pathology in French.
    \item \texttt{cond-name-eng}: name of the pathology in English.
    \item \texttt{icd10-id}: ICD-10 code of the pathology.
    \item \texttt{severity}: the severity associated with the pathology. The lower the more severe.
    \item \texttt{symptoms}: data structure describing the set of symptoms characterizing the pathology. Each symptom is represented by its corresponding \texttt{name} entry in the \texttt{release\_evidences.json} file.
    \item \texttt{antecedents}: data structure describing the set of antecedents characterizing the pathology. Each antecedent is represented by its corresponding \texttt{name} entry in the  \texttt{release\_evidences.json} file.
\end{itemize}

\paragraph{Patient description}
Each patient in each of the 3 sets has the following attributes:
\begin{itemize}
    \item \texttt{AGE}: the age of the synthesized patient.
    \item \texttt{SEX}: the sex of the synthesized patient.
    \item \texttt{PATHOLOGY}: name of the ground truth pathology (\texttt{condition\_name} property in the \texttt{release\_conditions.json} file) that the synthesized patient is suffering from.
    \item \texttt{EVIDENCES}: list of evidences experienced by the patient. An evidence can be either binary, categorical or multi-choice. A categorical or multi-choice evidence is represented in the format \texttt{[evidence-name]\_@\_[evidence-value]} where \texttt{[evidence-name]} is the name of the evidence (\texttt{name} entry in the \texttt{release\_evidences.json} file) and \texttt{[evidence-value]} is a value from the \texttt{possible-values} entry. Note that for a multi-choice evidence, it is possible to have several \texttt{[evidence-name]\_@\_[evidence-value]} items in the evidence list, with each item being associated with a different evidence value. A binary evidence is simply represented as \texttt{[evidence-name]}.
    \item \texttt{INITIAL\_EVIDENCE}: the evidence provided by the patient to kick-start an interaction with an ASD/AD system. This is useful during model evaluation for a fair comparison of ASD/AD systems as they will all begin an interaction with a given patient from the same starting point. The initial evidence is randomly selected from the binary evidences found in the  evidence list mentioned above (i.e., \texttt{EVIDENCES}) and it is part of this list.
    \item \texttt{DIFFERENTIAL\_DIAGNOSIS}: the ground truth differential diagnosis for the patient. It is represented as a list of pairs of the form \texttt{[[patho\_1, proba\_1], [patho\_2, proba\_2], $\cdots$]} where \texttt{patho\_i} is the pathology name (\texttt{condition\_name} entry in the \texttt{release\_conditions.json} file) and \texttt{proba\_i} is its related probability.
\end{itemize}

\subsection{General notes about the dataset's differentials}
\label{appx:general_notes_DDXPlus_ddx}

It is important to understand that the level of specificity, sensitivity and confidence that a physician will seek when evaluating a patient will be influenced by the clinical setting. The dataset was built for acute care and biased toward high mortality and morbidity pathologies. Physicians will tend to consider negative evidences as equally important in such a clinical context in order to evaluate high acuity diseases.

In the creation of the DDXPlus dataset, a small subset of the diseases was chosen to establish a baseline. Medical professionals have to consider this very important point when reviewing the results, as the differential is considerably smaller. A smaller differential means less potential evidences to collect. It is thus essential to understand this point when we look at the differential produced and the collected evidence.

\subsection{Responsibility statement}
The authors declare that they bear all responsibility for violations of rights related to this dataset.

\section{Dataset samples}
\label{app:patient_samples}

We showcase in this section some samples from the DDXPlus dataset. For each example, we show the age, sex, the geographical region, the ground truth pathology, the symptoms, the antecedents (past medical history), as well as the corresponding differential diagnosis of a synthetic patient. The first two samples include feedback provided by a doctor to explain the differential diagnosis.

\subsection*{Sample 1}

\begin{lstlisting}
Sex: M, Age: 47
Geographical region: North America
Pathology: PSVT
Symptoms:
---------
	 - I feel pain.
	 - The pain is:
		 * tugging
		 * burning
	 - The pain locations are:
		 * back of head
		 * top of the head
		 * temple(R)
	 - On a scale of 0-10, the pain intensity is 4
	 - On a scale of 0-10, the pain's location precision is 8
	 - On a scale of 0-10, the pace at which the pain appear is 5
	 - I feel like I am about to faint.
	 - I feel lightheaded and dizzy.
	 - I feel palpitations.
Antecedents:
------------
	 - I feel anxious.
	 - I regularly drink coffee or tea.
	 - I regularly consume energy drinks.
	 - I regularly take stimulant drugs.
	 - I have recently taken decongestants or substances that may have stimulant effects.
Differential diagnosis:
-----------------------
PSVT: 0.226, Anemia: 0.161, Panic attack: 0.142, Atrial fibrillation: 0.117, Anaphylaxis: 0.113, Cluster headache: 0.092, Chagas: 0.072, Scombroid food poisoning: 0.071, HIV (initial infection): 0.006
\end{lstlisting}

The differential is good, in the sense that we would need to ask additional questions on causes of palpitations, ranging from mood disorders, or search for causes for anemia. Although this is a patient with PSVT with many risk factors for it, a clinician will want to search for all the possibilities that would change the course of the care episode. It is important to note that the diagnosis space is significantly smaller than the frame of reference of a physician, hence doctors will tend to refine the medical history and seek to discriminate further in order to establish a final differential.

\subsection*{Sample 2}

\begin{lstlisting}
Sex: F, Age: 55
Geographical region: North America
Pathology: GERD
Symptoms:
---------
	 - I feel pain.
	 - The pain is:
		 * haunting
		 * tugging
		 * sickening
	 - The pain locations are:
		 * lower chest
		 * upper chest
		 * epigastric
	 - On a scale of 0-10, the pain intensity is 4
	 - The pain radiates to these locations:
		 * lower chest
		 * upper chest
	 - On a scale of 0-10, the pain's location precision is 6
	 - On a scale of 0-10, the pace at which the pain appear is 2
	 - I have a burning sensation that starts in my stomach then goes up into my throat, and can be associated with a bitter taste in my mouth.
	 - I am coughing.
	 - I have symptoms that get worse after eating.
	 - My symptoms worse when lying down and alleviated while sitting up.
Antecedents:
------------
	 - I am significantly overweight compared to people of the same height as me.
	 - I drink alcohol excessively.
	 - I smoke cigarettes.
	 - I have a hiatal hernia.
	 - I have had to use a bronchodilator in the past.
	 - I am pregnant.
Differential diagnosis:
-----------------------
GERD: 0.196, Bronchitis: 0.148, Pericarditis: 0.129, Spontaneous rib fracture: 0.098, Unstable angina: 0.098, Boerhaave: 0.093, Possible NSTEMI / STEMI: 0.068, Tuberculosis: 0.060, Stable angina: 0.055, Pancreatic neoplasm: 0.054
\end{lstlisting}

This is a good differential based on the positive features. This differential includes diseases we would further need to inquire about as most are very relevant to the patient’s demographic. In this example, we know that acute presentation of myocardial infarction in women can be atypical. Although in the differential, we would like to refine the presentation to better grasp the disease probabilities and tailor the tests to that risk level.

\subsection*{Sample 3}

\begin{lstlisting}
Sex: F, Age: 66
Geographical region: North America
Pathology: URTI
Symptoms:
---------
	 - I have had significantly increased sweating.
	 - I feel pain.
	 - The pain is:
		 * tedious
		 * sensitive
	 - The pain locations are:
		 * cheek(R)
		 * cheek(L)
		 * occiput
		 * temple(L)
	 - On a scale of 0-10, the pain intensity is 8
	 - On a scale of 0-10, the pain's location precision is 5
	 - On a scale of 0-10, the pace at which the pain appeared is 1
	 - I have fever.
	 - I have a sore throat.
	 - I have diffuse muscle pain.
	 - I am coughing.
Antecedents:
------------
	 - I have been in contact with a person with similar symptoms in the past 2 weeks.
	 - I live with 4 or more people.
	 - I smoke cigarettes.
	 - I am exposed to secondhand cigarette smoke on a daily basis.
Differential diagnosis:
-----------------------
URTI: 0.208, Influenza: 0.204, Bronchitis: 0.186, Pneumonia: 0.146, Tuberculosis: 0.121, HIV (initial infection): 0.078, Chagas: 0.057

\end{lstlisting}

\subsection*{Sample 4}

\begin{lstlisting}
Sex: M, Age: 69
Geographical region: North America
Pathology: Anaphylaxis
Symptoms:
---------
	 - I have eaten something that I have an allergy to.
	 - I feel pain.
	 - The pain is:
		 * a knife stroke
		 * sharp
	 - The pain locations are:
		 * flank(L)
		 * pubis
		 * belly
		 * epigastric
	 - On a scale of 0-10, the pain intensity is 6
	 - On a scale of 0-10, the pace at which the pain appeared is 4
	 - I have rashes.
	 - The color of the rash is pink
	 - On a scale of 0-10, the rash swelling is 6
	 - The regions affected by the rash are:
		 * biceps(R)
		 * biceps(L)
		 * mouth
		 * thyroid cartilage
		 * ankle(R)
	 - On a scale of 0-10, the pain intensity caused by the rash is 2
	 - The rash lesions are larger than 1cm
	 - On a scale of 0-10, the severity of the rash itching is 10
	 - I have nausea.
	 - I have swelling in one or more areas of my body.
	 - The swelling locations are:
		 * forehead
		 * cheek(R)
		 * cheek(L)
		 * nose
	 - I have noticed a high pitched sound when breathing in.
Antecedents:
------------
	 - I have a known severe food allergy.
	 - I am more likely to develop common allergies than the general population.
Differential diagnosis:
-----------------------
Anaphylaxis: 0.159, Possible NSTEMI / STEMI: 0.112, Localized edema: 0.110, GERD: 0.110, Unstable angina: 0.104, Larygospasm: 0.095, Boerhaave: 0.093, Pulmonary embolism: 0.069, Chagas: 0.057, Pericarditis: 0.046, Stable angina: 0.045
\end{lstlisting}

\subsection*{Sample 5}

\begin{lstlisting}
Sex: F, Age: 36
Geographical region: North America
Pathology: Pulmonary embolism
Symptoms:
---------
	 - I feel pain.
	 - The pain is:
		 * a knife stroke
	 - The pain locations are:
		 * side of the chest(L)
		 * breast(R)
		 * breast(L)
		 * posterior chest wall(R)
		 * posterior chest wall(L)
	 - On a scale of 0-10, the pain intensity is 1
	 - The pain radiates to these locations:
		 * side of the chest(L)
		 * scapula(R)
		 * scapula(L)
		 * posterior chest wall(R)
		 * posterior chest wall(L)
	 - On a scale of 0-10, the pain's location precision is 3
	 - On a scale of 0-10, the pace at which the pain appeared is 9
	 - I am experiencing shortness of breath or difficulty breathing in a significant way.
	 - I have swelling in one or more areas of my body.
	 - The swelling locations are:
		 * calf(L)
	 - I have pain which increases when I breathe in deeply.
Antecedents:
------------
Differential diagnosis:
-----------------------
Pulmonary embolism: 0.091, Panic attack: 0.072, Anaphylaxis: 0.071, Spontaneous pneumothorax: 0.070, Possible NSTEMI / STEMI: 0.069, Unstable angina: 0.066, Stable angina: 0.054, Pulmonary neoplasm: 0.054, Pericarditis: 0.053, Guillain-Barre syndrome: 0.051, Myocarditis: 0.050, Atrial fibrillation: 0.049, Acute pulmonary edema: 0.044, SLE: 0.041, Acute dystonic reactions: 0.036, Myasthenia gravis: 0.036, Anemia: 0.036, Chagas: 0.030, Sarcoidosis: 0.029
\end{lstlisting}

\subsection*{Sample 6}

\begin{lstlisting}
Sex: M, Age: 43
Geographical region: North America
Pathology: Panic attack
Symptoms:
---------
	 - I feel pain.
	 - The pain is:
		 * a cramp
	 - The pain locations are:
		 * side of the chest(R)
		 * flank(R)
		 * hypochondrium(R)
		 * hypochondrium(L)
		 * belly
	 - On a scale of 0-10, the pain intensity is 6
	 - On a scale of 0-10, the pain's location precision is 9
	 - On a scale of 0-10, the pace at which the pain appeared is 7
	 - I am experiencing shortness of breath or difficulty breathing in a significant way.
	 - I feel lightheaded and dizzy.
	 - I feel like I am dying.
	 - I have nausea.
	 - I feel palpitations.
	 - I feel like I am detached from my own body or my surroundings.
	 - I have recently had numbness, loss of sensation or tingling on my body.
Antecedents:
------------
	 - Some members of my immediate family have a psychiatric illness.
	 - I drink alcohol excessively.
	 - I have been diagnosed with depression.
	 - I suffer from chronic anxiety.
	 - Some family members are known to have migraines.
	 - I suffer from fibromyalgia.
	 - I have had a head trauma.
Differential diagnosis:
-----------------------
Panic attack: 0.094, PSVT: 0.076, Possible NSTEMI / STEMI: 0.073, Spontaneous pneumothorax: 0.066, Anaphylaxis: 0.063, Unstable angina: 0.061, Anemia: 0.059, Guillain-Barre syndrome: 0.058, Boerhaave: 0.056, Atrial fibrillation: 0.054, Acute pulmonary edema: 0.053, Pulmonary embolism: 0.049, Scombroid food poisoning: 0.044, GERD: 0.041, Stable angina: 0.038, Acute dystonic reactions: 0.033, Myasthenia gravis: 0.033, Sarcoidosis: 0.027, Chagas: 0.025
\end{lstlisting}

\subsection*{Sample 7}

\begin{lstlisting}
Sex: F, Age: 63
Geographical region: North America
Pathology: Bronchitis
Symptoms:
---------
	 - I feel pain.
	 - The pain is:
		 * burning
	 - The pain locations are:
		 * lower chest
		 * side of the chest(L)
		 * pharynx
	 - On a scale of 0-10, the pain intensity is 3
	 - On a scale of 0-10, the pain's location precision is 2
	 - On a scale of 0-10, the pace at which the pain appeared is 3
	 - I am experiencing shortness of breath or difficulty breathing in a significant way.
	 - My cough produces colored or more abundant sputum than usual.
	 - I have fever.
	 - I have a sore throat.
	 - I am coughing.
	 - I have noticed a wheezing sound when I exhale.
Antecedents:
------------
	 - I have a chronic obstructive pulmonary disease.
Differential diagnosis:
-----------------------
Acute COPD exacerbation / infection: 0.075, Pneumonia: 0.070, Bronchitis: 0.070, Bronchiectasis: 0.065, Panic attack: 0.053, Pulmonary neoplasm: 0.052, Tuberculosis: 0.052, Possible NSTEMI / STEMI: 0.051, GERD: 0.051, Unstable angina: 0.049, Pericarditis: 0.049, URTI: 0.045, Boerhaave: 0.045, Stable angina: 0.041, Acute laryngitis: 0.039, Atrial fibrillation: 0.036, Viral pharyngitis: 0.035, Guillain-Barre syndrome: 0.026, Acute dystonic reactions: 0.026, Myocarditis: 0.026, Sarcoidosis: 0.021, PSVT: 0.014, Influenza: 0.005, Chagas: 0.005
\end{lstlisting}

\subsection*{Sample 8}

\begin{lstlisting}
Sex: F, Age: 35
Geographical region: North America
Pathology: Acute rhinosinusitis
Symptoms:
---------
	 - I feel pain.
	 - The pain is:
		 * burning
		 * sharp
	 - The pain locations are:
		 * forehead
		 * cheek(R)
		 * nose
		 * eye(L)
	 - On a scale of 0-10, the pain intensity is 2
	 - The pain radiates to these locations:
		 * forehead
		 * nose
	 - On a scale of 0-10, the pain's location precision is 8
	 - I have fever.
	 - I have lost my sense of smell.
	 - I have greenish/yellowish nasal discharge.
	 - I am coughing.
Antecedents:
------------
	 - I have had a cold in the last 2 weeks.
	 - I have had to use a bronchodilator in the past.
	 - I have been diagnosed with gastroesophageal reflux.
	 - my vaccinations are up to date.
	 - I am more likely to develop common allergies than the general population.
Differential diagnosis:
-----------------------
Acute rhinosinusitis: 0.232, Chronic rhinosinusitis: 0.204, Bronchitis: 0.198, Tuberculosis: 0.146, Influenza: 0.145, Chagas: 0.064, Pneumonia: 0.011
\end{lstlisting}

\subsection*{Sample 9}

\begin{lstlisting}
Sex: M, Age: 49
Geographical region: North America
Pathology: Allergic sinusitis
Symptoms:
---------
	 - My nose or the back of my throat is itchy.
	 - I have severe itching in one or both eyes.
	 - I have nasal congestion.
	 - I am coughing.
Antecedents:
------------
	 - Some family members suffer from allergies, hay fever or eczema.
	 - Some family members have asthma.
Differential diagnosis:
-----------------------
Allergic sinusitis: 0.348, Bronchitis: 0.251, URTI: 0.201, Influenza: 0.200
\end{lstlisting}

\subsection*{Sample 10}

\begin{lstlisting}
Sex: M, Age: 32
Geographical region: North America
Pathology: Pulmonary embolism
Symptoms:
---------
	 - I have been coughing up blood.
	 - I feel pain.
	 - The pain is:
		 * a knife stroke
		 * sharp
	 - The pain locations are:
		 * upper chest
		 * breast(R)
		 * breast(L)
		 * posterior chest wall(R)
		 * posterior chest wall(L)
	 - On a scale of 0-10, the pain intensity is 10
	 - The pain radiates to these locations:
		 * side of the chest(R)
		 * side of the chest(L)
		 * breast(L)
		 * posterior chest wall(R)
		 * posterior chest wall(L)
	 - On a scale of 0-10, the pain's location precision is 9
	 - On a scale of 0-10, the pace at which the pain appeared is 8
	 - I am experiencing shortness of breath or difficulty breathing in a significant way.
	 - I lost consciousness.
	 - I have pain which increases when I breathe in deeply.
Antecedents:
------------
	 - I have had surgery within the last month.
Differential diagnosis:
-----------------------
Pulmonary embolism: 0.098, Pericarditis: 0.080, Spontaneous pneumothorax: 0.079, Possible NSTEMI / STEMI: 0.074, Panic attack: 0.072, Unstable angina: 0.066, Boerhaave: 0.061, Acute pulmonary edema: 0.058, Myocarditis: 0.057, Stable angina: 0.054, Guillain-Barre syndrome: 0.051, Atrial fibrillation: 0.049, GERD: 0.046, Acute dystonic reactions: 0.036, Myasthenia gravis: 0.036, Anemia: 0.036, Sarcoidosis: 0.029, PSVT: 0.019
\end{lstlisting}

\subsection*{Sample 11}

\begin{lstlisting}
Sex: M, Age: 75
Geographical region: North America
Pathology: Pericarditis
Symptoms:
---------
	 - I feel pain.
	 - The pain is:
		 * sharp
	 - The pain locations are:
		 * upper chest
		 * breast(R)
		 * breast(L)
	 - On a scale of 0-10, the pain intensity is 7
	 - The pain radiates to these locations:
		 * posterior chest wall(L)
	 - On a scale of 0-10, the pain's location precision is 7
	 - On a scale of 0-10, the pace at which the pain appeared is 7
	 - I feel palpitations.
	 - My symptoms worse when lying down and alleviated while sitting up.
	 - I have pain which increases when I breathe in deeply.
Antecedents:
------------
	 - I have had a pericarditis.
Differential diagnosis:
-----------------------
Pericarditis: 0.180, Pulmonary embolism: 0.112, Panic attack: 0.111, Possible NSTEMI / STEMI: 0.111, Unstable angina: 0.104, Boerhaave: 0.103, PSVT: 0.098, Stable angina: 0.054, GERD: 0.050, Scombroid food poisoning: 0.042, Spontaneous pneumothorax: 0.035
\end{lstlisting}

\section{Model Analysis}
\label{appx:model_analysis}

We present in this section the sequence of question-answer pairs as well as the differentials predicted by AARLC and BASD for the patient introduced in Section~\ref{sec:datanalysis}. Two variants are considered for each model, one trained to predict the ground truth differential and one trained to predict the ground truth pathology. At the beginning of the interaction with a model, the patient provides her age, sex, and an initial evidence. The behavior of all 4 models is evaluated by a doctor.

\subsection*{AARLC trained to predict the differential diagnosis}

\begin{lstlisting}
Sex: F, Age: 79
Initial evidence:
-----------------
    I have symptoms that increase with physical exertion but alleviate with rest

Agent inquiries:
----------------
- Characterize your pain:
	 * a knife stroke
- Do you have pain somewhere, related to your reason for consulting?
	 Y
- How fast did the pain appear?
	 * 9
- How precisely is the pain located?
	 * 4
- How intense is the pain?
	 * 7
- Where is the affected region located?
	 nowhere
- Do you feel pain somewhere?
	 * upper chest
	 * breast(R)
	 * breast(L)
- Does the pain radiate to another location?
	 * nowhere
- Where is the swelling located?
	 nowhere
- Have you traveled out of the country in the last 4 weeks?
	 * N
- Are you experiencing shortness of breath or difficulty breathing in a significant way?
	 N
- Do you have a cough?
	 N
- Do you have a chronic obstructive pulmonary disease (COPD)?
	 Y
- Do you have a fever (either felt or measured with a thermometer)?
	 N
- Do you feel lightheaded and dizzy or do you feel like you are about to faint?
	 N
- Do you smoke cigarettes?
	 Y
- Have you ever had a spontaneous pneumothorax?
	 Y

Predicted Differential:
-----------------------
Unstable angina, Stable angina, Possible NSTEMI / STEMI, Spontaneous pneumothorax, GERD, Pericarditis, Pulmonary embolism, Atrial fibrillation

\end{lstlisting}

\subsection*{AARLC trained to predict the ground truth pathology}
\begin{lstlisting}
Sex: F, Age: 79
Initial evidence:
-----------------
    I have symptoms that increase with physical exertion but alleviate with rest

Agent inquiries:
----------------
- Do you have pain somewhere, related to your reason for consulting?
	 Y
- Does the pain radiate to another location?
	 * nowhere
- Where is the affected region located?
	 nowhere
- Do you feel pain somewhere?
	 * upper chest
	 * breast(R)
	 * breast(L)
- How intense is the pain?
	 * 7

Predicted Differential:
-----------------------
Myocarditis, Spontaneous pneumothorax

\end{lstlisting}

\subsection*{BASD trained to predict the differential diagnosis}
\begin{lstlisting}
Sex: F, Age: 79
Initial evidence:
-----------------
    I have symptoms that increase with physical exertion but alleviate with rest

Agent inquiries:
----------------
- Do you have pain somewhere, related to your reason for consulting?
	 Y
- Do you feel pain somewhere?
	 * upper chest
	 * breast(R)
	 * breast(L)
- How intense is the pain?
	 * 7
- Characterize your pain:
	 * a knife stroke
- How precisely is the pain located?
	 * 4
- How fast did the pain appear?
	 * 9
- Do you smoke cigarettes?
	 Y
- Have you ever had a spontaneous pneumothorax?
	 Y
- Do you have a chronic obstructive pulmonary disease (COPD)?
	 Y

Predicted Differential:
-----------------------
Unstable angina, Possible NSTEMI / STEMI, Stable angina, Pericarditis, Atrial fibrillation, Spontaneous pneumothorax, GERD

\end{lstlisting}

\subsection*{BASD trained to predict the ground truth pathology}
\begin{lstlisting}
Sex: F, Age: 79
Initial evidence:
-----------------
    I have symptoms that increase with physical exertion but alleviate with rest

Agent inquiries:
----------------
- Characterize your pain:
	 * a knife stroke
- How intense is the pain?
	 * 7
- Do you have pain somewhere, related to your reason for consulting?
	 Y
- How precisely is the pain located?
	 * 4
- Do you feel pain somewhere?
	 * upper chest
	 * breast(R)
	 * breast(L)
- How fast did the pain appear?
	 * 9
- Are you experiencing shortness of breath or difficulty breathing in a significant way?
	 N
- Do you have pain that is increased when you breathe in deeply?
	 N
- Are your symptoms worse when lying down and alleviated while sitting up?
	 N
- Do you smoke cigarettes?
	 Y
- Do you have a chronic obstructive pulmonary disease (COPD)?
	 Y
- Have you ever had a spontaneous pneumothorax?
	 Y
- Have any of your family members ever had a pneumothorax?
	 Y
- Do you have chest pain even at rest?
	 Y

Predicted Differential:
-----------------------
Spontaneous pneumothorax

\end{lstlisting}

Following is the evaluation performed by the doctor:
\begin{itemize}
\item AARLC trained on the ground truth pathology is the worst, having too few questions and an incomplete differential.

\item BASD trained on the differential has a good differential, but too few collected evidences, making the collected evidence unspecific. More questions are needed to adequately cover the proposed differential.

\item AARLC trained with the differential and BASD trained on the ground truth pathology produce the best results, gather a sensible amount of evidence. AARLC actively searches a wider array of diseases, searching also for infectious causes. BASD is still very good,  unsurprisingly searching for a pathology confirmation and goes a bit further to ask specific questions. A clinician would like the best of both worlds, but AARLC covers a wider range of diseases, although a doctor would need to complete the evaluation with more questions.

Pertinent questions will be influenced by the clinical context. In an acute care setting, a physician will try to gain specificity and include more questions centered around high acuity diseases in order to wisely choose the following medical tests to rule-in or rule-out the diseases. 

\end{itemize}

\end{document}